\documentclass[runningheads]{llncs}

\usepackage{eccv}

\usepackage{eccvabbrv}

\usepackage{graphicx}
\usepackage{booktabs}

\usepackage[accsupp]{axessibility}  

\usepackage{hyperref}

\usepackage{orcidlink}
\usepackage{multirow}

\newcommand{\benchmark}[1]{GSI-Bench}

\usepackage[most]{tcolorbox}
\definecolor{takeaway}{rgb}{0.97,0.94,0.91}
\newtcbox{\takeawaybox}{on line,
  colback=takeaway,
  colframe=takeaway,
  boxrule=0pt,
  arc=2pt,       
  boxsep=1pt,    
  left=3pt,right=3pt,top=1pt,bottom=1pt
}
\newtcolorbox{takeawayblock}{
  colback=takeaway,
  colframe=takeaway,
  boxrule=0pt,
  arc=2pt,
  left=2pt,right=2pt,top=2pt,bottom=2pt,
  boxsep=0pt,
  before skip=1pt,
  after skip=1pt
}

\begin{document}

\title{Imaginative Perception Tokens Enhance Spatial Reasoning in Multimodal Language Models} 

\titlerunning{Imaginative Perception Tokens}

\author{Mahtab Bigverdi\inst{1,2*} \and
Linjie Li\inst{1*} \and
Weikai Huang\inst{1,2*} \and Yiming Liu\inst{1} \and \\ Jaemin Cho\inst{1,2,5}  \and  Tuhin Kundu\inst{3} \and Chris Dongjoo Kim\inst{2} \and  Zelun Luo\inst{4} \and \\  Jieyu Zhang\inst{1,2} \and   Linda Shapiro\inst{1} \and  Ranjay Krishna\inst{1}}

\authorrunning{Bigverdi et al.}

\institute{University of Washington, Seattle, WA, USA \and Allen Institute for AI, Seattle, WA, USA \and Microsoft, Seattle, WA, USA \and OpenAI, San Francisco, CA, USA \and John Hopkins University, Baltimore, Maryland, USA\\
\email{\{mahtab,linjli,weikaih\}@cs.washington.edu}}

\maketitle

\begin{abstract}
Vision-language models (VLMs) excel at many tasks, yet continue to struggle with spatial reasoning, problems where the key information is not directly observable in the input. Many spatial questions require \emph{imaginative perception}: simulating an unseen viewpoint, tracing a trajectory through an occluded space, or integrating partial views into a coherent spatial map. Humans naturally support this kind of reasoning through imagination. Prior work has introduced intermediate visual representations (e.g., visual thoughts, depth, or box tokens), but these intermediates often refine structure already visible rather than predicting the missing spatial structure implied by the evidence.
We introduce \textbf{Imaginative Perception Tokens (IPT)}, intermediate perceptual representations that externalize what a VLM would perceive under an alternative spatial configuration while remaining consistent with the observed input. To study this capability, we formulate three tasks that require imaginative perception: \textbf{Perspective Taking (PET)}, \textbf{Path Tracing (PT)}, and \textbf{Multiview Counting (MVC)}. For each task, we construct datasets of $\sim$20K examples spanning simulated and real-world settings, paired with ground-truth intermediate imaginations, final answers, and curated evaluation benchmarks.
Using the unified VLM BAGEL~\cite{deng2025bagel} as our backbone, IPT supervision improves spatial reasoning across several settings and often outperforms textual chain-of-thought training, even when no image is generated at inference time. For example, on MVC, IPT improves accuracy by 3.4\% and achieves performance competitive with strong closed-source models on Path Tracing. We also find that mixed training with IPT and label-only data can further improve performance. In contrast, textual chain-of-thought can be detrimental on these tasks, substantially degrading performance in some cases, highlighting a modality mismatch when forcing spatial computation through language. Overall, IPT provides a principled supervision signal for reasoning over unobserved structure, yielding stronger spatial generalization and a more interpretable intermediate aligned with the underlying geometry of the task. Code will be released at the \href{https://mahtabbigverdi.github.io/Imaginative-tokens.github.io/}{project page}.


\end{abstract}
\begin{figure}[t]
    \centering
    \includegraphics[width=0.95\linewidth]{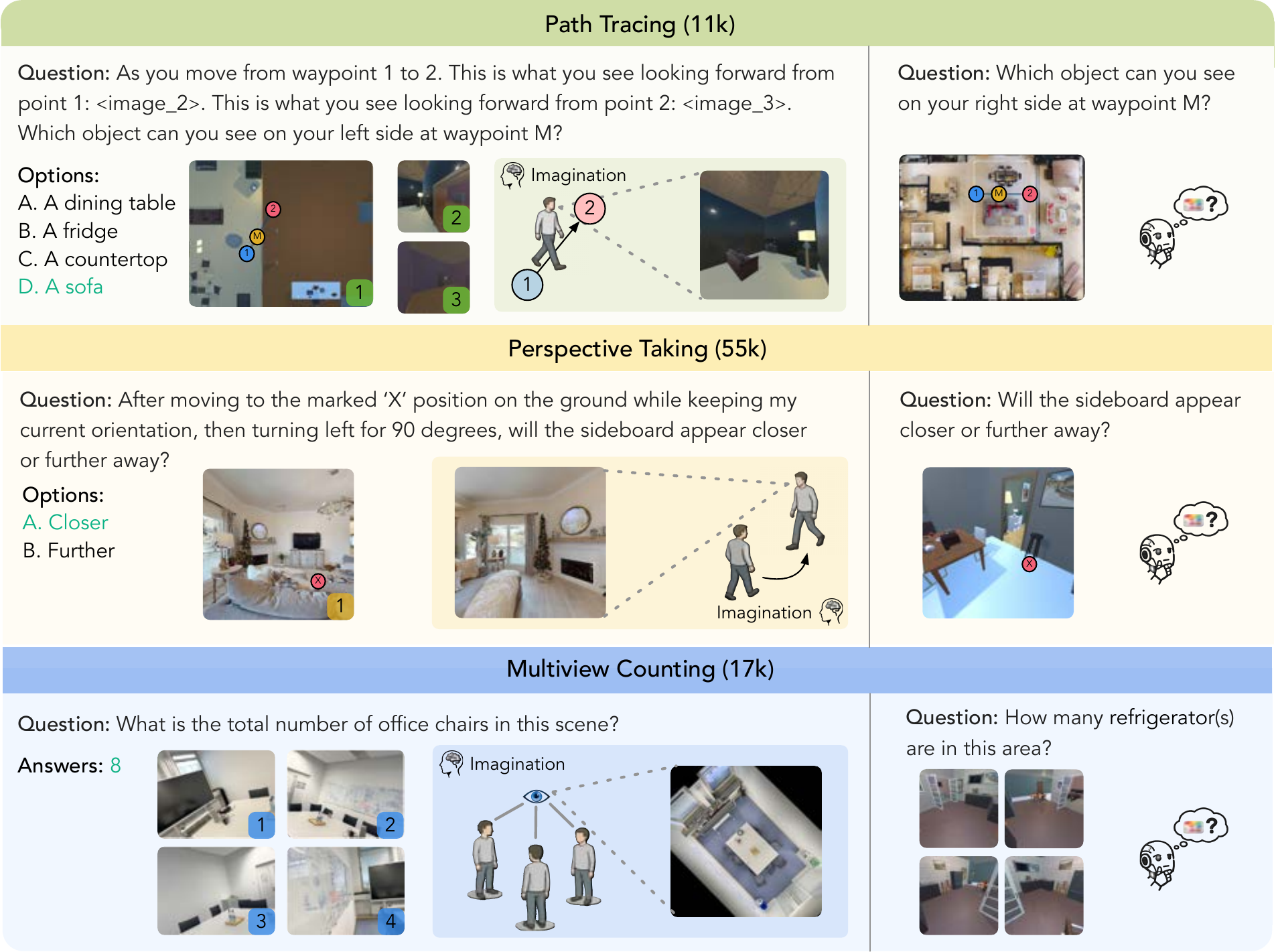}
    \caption{\textbf{Overview of the three spatial imagination tasks.} The left columns show training examples with ground-truth imaginative perception; the right columns show evaluation examples.}
    \label{fig:data}
\end{figure}

\section{Introduction}
\label{sec:intro}

Spatial reasoning remains a persistent challenge for vision-language models (VLMs)~\cite{deitke2025molmo,clark2026molmo2,bai2025qwen3vl}. These models struggle particularly with tasks that require reasoning about 3D spatial relationships, how objects relate within three-dimensional environments, how these relationships transform under viewpoint changes, or how to integrate information from multiple partial observations into a coherent scene representation~\cite{yin2025mindcube, kamath2023whatsup}. While current VLMs can recognize objects and attributes effectively, they frequently fail when spatial reasoning demands manipulating structure, such as predicting scene appearance from novel viewpoints~\cite{li2025viewspatialbench, ma20253dsrbench} or aggregating information across multiple views~\cite{yang2025mmsibench}.

A key reason for this difficulty is that many spatial reasoning problems cannot be solved by analyzing the input alone. Instead, they require constructing a spatial representation that is not directly observed. Humans naturally address such problems through imagination: when asked what lies to the left after moving to a new position, or how many objects exist in a room seen from several viewpoints, we mentally simulate the scene from unseen perspectives or integrate partial observations into a unified spatial map~\cite{yang2025thinkinginspace,yin2025mindcube,zhang2026theoryofspace}. In other words, spatial reasoning often depends on imagining missing spatial structure that proceed despite incomplete observations.

Existing approaches provide only partial solutions. Recent work teaches models to generate intermediate visual thoughts alongside language~\cite{gu2025thinkmorph, li2025mvot, hu2024visualsketchpad}, while others introduce structured perceptual intermediates, such as depth maps or bounding boxes represented as tokens~\cite{bigverdi2024aurora, yang2025mirage, ray2025mulltokens}. Although these methods demonstrate that intermediate visual representations can support reasoning, they primarily operate over information already present in the input observation, refining visible structures or extracting perceptual attributes. However, as discussed above, many spatial reasoning problems arise precisely because the required spatial information is not directly observable, and therefore requires imagination.

To address this gap, we propose \textbf{Imaginative Perception Tokens} for VLMs. When VLMs are trained with them, they enable intermediate reasoning steps that represent novel spatial views. Unlike standard perceptual intermediates that describe structures visible in the input, imaginative representations correspond to what the model would perceive if it were observing the input from a different spatial configuration, such as from an unseen viewpoint or after integrating multiple partial observations into one. At the same time, they are not unconstrained imagination: the predicted percept must remain consistent with the observed scene. These tokens externalize the model's prediction of what would be perceived given incomplete spatial evidence. 

To study this capability, we propose three spatial reasoning tasks that fundamentally require imaginative perception. (1) Perspective Taking requires predicting how a scene would appear from a new viewpoint given a single first-person observation (“If you move to the marked position and turn left, will the chair appear on your left or right?"); (2) Path Tracing requires inferring what an agent would see along a navigation path based on a top-down view (“If you walk along the marked path, which object will you see on your side?”); Finally, (3) Multiview Counting requires integrating multiple partial observations into a top-down view to determine the number of objects present in the scene. These tasks would be made easy when correctly predicting what would be perceived in a different spatial configuration. 
For each task we construct a dataset of approximately 20k examples each drawn from both real-world and synthetic simulated environments, with ground-truth intermediate spatial imaginations paired with final answers. Each dataset is accompanied by a human-filtered benchmark for evaluation. Together these constitute the first datasets designed explicitly to train and evaluate visually-grounded intermediate spatial reasoning in models.

Empirically, we find that training with imaginative perception supervision can improve performance on these spatial reasoning tasks compared to answer-only supervision, and often compares favorably to textual chain-of-thought approaches. These improvements can persist even when the model does not explicitly generate intermediate images at inference time, suggesting that such supervision may help models develop stronger internal spatial representations. At the same time, we observe that the benefits vary across tasks and settings, indicating that imagination quality and task structure both play important roles.

Overall, our results suggest that supervising models with intermediate perceptual predictions offers a useful direction for improving spatial reasoning, particularly in settings where the required structure is not directly observable from the input.

\section{Related Works}
\label{sec:rw}

\noindent\textbf{Evaluation of VLMs' spatial reasoning.}
A growing body of benchmarks has established that spatial reasoning remains a persistent weakness of modern vision-language models. Early datasets target brittleness in basic spatial predicates: SpatialSense~\cite{yang2019spatialsense} reduces language priors through adversarial crowdsourcing, while VSR~\cite{liu2022vsr} scales relation types in a caption-verification format, and What'sUp~\cite{kamath2023whatsup} uses minimal-pair testing to reveal systematic failures on left/right and above/below distinctions. More recent work shifts from 2D relations to viewpoint and 3D structure. 3DSRBench~\cite{ma20253dsrbench} shows that models fail under modest changes in perspective, depth, and occlusion, and ViewSpatial-Bench~\cite{li2025viewspatialbench} identifies a ``perspective gap'': models often succeed in camera-centered views but break when asked to adopt human-centered viewpoints.

Benchmarks have also expanded to multi-image and video settings where maintaining a consistent spatial state is essential. VSI-Bench~\cite{yang2025thinkinginspace} tests whether models can build a persistent mental map from videos, while MMSI-Bench~\cite{yang2025mmsibench} reports large human--model gaps on cross-view scene reconstruction. MindCube~\cite{yin2025mindcube} is closely aligned with our motivation, targeting spatial mental modeling from limited views, including perspective-taking and ``what-if'' scene dynamics. Counting Stacked Objects~\cite{dumery2025countingstackedobjects} studies 3D object counting under heavy occlusion across multiple views, directly analogous to our Multiview Counting setting. Finally, benchmark design work emphasizes that shortcuts remain pervasive: Brown \etal~\cite{brown2025trainontestset} construct VSI-Bench-Debiased by iteratively pruning samples solvable via priors, reinforcing the need for evaluations where success requires genuine spatial computation.

Collectively, these benchmarks diagnose \emph{where} VLMs fail spatially, but they typically evaluate \emph{discriminative} understanding---reading off a relation from an observed view---rather than \emph{constructive} spatial imagination. Our work is complementary: we isolate \emph{imaginative perception} as a standardized intermediate substrate, and pair each task with a ground-truth intermediate spatial imagination rather than only a final answer label.

\noindent\textbf{Intermediate representations for spatial reasoning.}
Chain-of-thought prompting~\cite{wei2022cot} can improve multi-step reasoning, but serializing viewpoint transformations, occlusions, and geometric constraints into language is often awkward and error-prone. This motivates intermediate representations in modalities better aligned with spatial computation. One direction externalizes reasoning into explicit visual buffers: Visual Sketchpad~\cite{hu2024visualsketchpad} equips models with drawing actions for iterative refinement, and MVoT~\cite{li2025mvot} introduces visualization-of-thought traces that help on dynamic spatial tasks where text CoT struggles. ThinkMorph~\cite{gu2025thinkmorph} studies interleaved text--image reasoning traces, and OpenAI describes o3/o4-mini as using chains-of-thought that include simple image transformations during reasoning~\cite{openai2025thinkingimages}. A complementary line introduces latent visual scratchpads: Mirage~\cite{yang2025mirage} frames latent tokens as ``machine mental imagery,'' and Mull-Tokens~\cite{ray2025mulltokens} generalizes to modality-agnostic latent thinking tokens.

Our work differs in what the intermediate is meant to represent. Many prior approaches treat intermediate images or latents as optional visualizations of \emph{visible} structure. We instead target \emph{imaginative perception}: predicting what would be perceived under an unobserved spatial configuration (e.g., a rotated viewpoint or a top-down path state), a representation constrained by the input but not present in it. This framing provides a principled criterion for when intermediate visual thoughts are necessary and a controlled way to supervise them.

\noindent\textbf{Unified multimodal models for interleaved understanding and generation.}
Producing imaginative perceptual intermediates within a single model requires the ability to both understand and generate images. Unified decoder-only architectures treat image tokens as first-class sequence elements, enabling arbitrary text--image interleaving. Chameleon~\cite{chameleon2024chameleon} is an early example, while Show-o2~\cite{xie2025showo2} and Janus~\cite{chen2025januspro} offer alternative unified designs that balance understanding and generation. We build on BAGEL~\cite{deng2025bagel}, a unified model pretrained on large interleaved corpora that exhibits strong spatial capabilities, making it a natural substrate for producing intermediate spatial imaginations. Crucially, however, a unified architecture alone does not guarantee that intermediate images are \emph{used} in a way that supports reasoning; our work provides task constructions and supervision that make imaginative perception the relevant computational substrate.
\section{Spatial Imagination: Tasks and Datasets}
\label{sec:data}


We introduce three spatial reasoning tasks that require constructing a missing spatial representation from incomplete inputs (single-view, partial-view, or map inputs).
For each task, we build a $10$k--$50$k training set with paired \emph{ground-truth spatial imaginations} (task-specific intermediate visual supervision) and final answers, and we release a human-filtered benchmark for controlled evaluation. All datasets will be released publicly; for consistency across tasks, we train our models on the AI2-THOR~\cite{kolve2017ai2} subset of each training set. 
Table~\ref{tab:dataset_stats} and Fig~\ref{fig:data} summarize the training data and evaluation benchmarks.


\begin{table}[t]
  \caption{\textbf{Dataset and benchmark statistics.} All experiments use the AI2-THOR subset for training; additional data sources are released for future research. $\dagger$: human-verified subset.}
  \vspace{-1em}
  \centering
  \resizebox{\columnwidth}{!}{
    \begin{tabular}{l l l}
    \toprule
    \textbf{Task} &  \\
    \midrule
    \multirow{4}{*}{\textbf{Perspective Taking}}
      & Source (samples) & AI2-THOR (20,531) + Habitat (19,998) + Real images (15,000) \\
      & Train            & 55,529 \\
      & Eval             & AI2-THOR$^\dagger$ (238), Habitat$^\dagger$ (300) \\
      & IPT Format & Novel-viewpoint image \\
    \midrule
    \multirow{4}{*}{\textbf{Path Tracing}}
      & Source (samples) & AI2-THOR (11,204) \\
      & Train            & 11,204 \\
      & Eval             & AI2-THOR$^\dagger$ (329), Real$^\dagger$ (332) \\
      & IPT Format & Sideview image \\
    \midrule
    \multirow{4}{*}{\textbf{Multiview Counting}}
      & Source (samples) & AI2-THOR (17,079) + MessyTable (1,880) + ScanNet (540) \\
      & Train            & 19,499 \\
      & Eval             & AI2-THOR$^\dagger$ (260) \\
      & IPT Format & Top-down BEV map \\
    \bottomrule
    \end{tabular}
  }
  \label{tab:dataset_stats}
\end{table}

\subsection{Perspective Taking}
\label{sec:perspective_taking}

Given a first-person view of an indoor scene with target positions marked, the model must answer a spatial question (\eg, ``After moving to `X' and turning left 90°, will the \{object\} be on your left or right?'') about the scene from the new viewpoint. Since the target view is never provided, the model must mentally simulate the spatial transformation rather than read off the answer directly.

\noindent\textbf{Sub-categories.}
Questions span two spatial relation types across six balanced sub-categories.
\emph{Distance change} asks whether a target object becomes closer or further after the viewpoint shift: (1)~\emph{closer} and (2)~\emph{further}.
\emph{Relative position} asks whether the object falls to the left or right in the new view, defined by the object's lateral position before and after the transformation: (3)~\emph{left$\to$left}; (4)~\emph{left$\to$right}; (5)~\emph{right$\to$left}; (6)~\emph{right$\to$right}.
Overall accuracy is the unweighted mean across all six sub-categories so that each spatial relationship contributes equally, preventing models from gaming the metric by over-predicting common cases.

\noindent\textbf{Imaginative perception target.} A novel-viewpoint rendering of the scene from the target position, directly supervised against ground-truth renders from the 3D scene.

\noindent\textbf{Data.}
Synthetic data is generated from AI2-THOR~\cite{kolve2017ai2} and Habitat~\cite{habitat19iccv,szot2021habitat,puig2023habitat3} by sampling source/target camera pairs, rendering first-person views, and annotating the source view with a red ``X'' marking the target.
Questions cover two relation types (distance change, relative position) across six balanced sub-categories.
A \emph{mixed} training data variant additionally incorporates real-world examples from the Visual Spatial Tuning dataset~\cite{yang2025visual} (camera motion subset) as a synthetic-to-real bridge.
The base training set contains $20{,}531$ AI2-THOR examples; the mixed variant totals $55{,}529$.
We evaluate on held-out human-verified AI2-THOR ($238$) and Habitat ($300$) benchmarks.
Full sub-category breakdowns and data generation details are provided in the Appendix.

\subsection{Path Tracing}
\label{sec:path_tracing}

Given a top-down map with a marked path $1\!\rightarrow\!2$, a midpoint $M_1$, and egocentric forward views at waypoints 1 and 2, the model must identify which object is visible on a queried side at $M_1$. Neither the top-down map nor the endpoint views reveal first-person visibility at the midpoint, requiring the model to imagine what the agent would see from ground level.

We evaluate under three input settings of increasing spatial cues: \emph{Path} (map only), \emph{PathArr} (map + query direction arrow), and \emph{EgoDir} (map + egocentric endpoint views).

\noindent\textbf{Imaginative perception target.} A sideview image — a first-person rendering from $M_1$ — that externalizes the 3D visibility reasoning the top-down input cannot support. Ground-truth sideviews are rendered directly from the simulator at $M_1$.

\noindent\textbf{Data.}
Synthetic data is generated from AI2-THOR~\cite{kolve2017ai2} and ProcTHOR~\cite{deitke2022procthor}, sampling feasible two-waypoint paths balanced across room types and distance bins.
Questions are template-generated with four answer choices and quality-filtered via TIFA-style verification~\cite{hu2023tifa} using GPT-4.1 majority voting; samples answerable from endpoint views alone are removed to ensure genuine imagination is required.
The synthetic training set contains $11{,}204$ examples.
A real-world benchmark of $332$ human-verified questions is constructed from Matterport3D~\cite{chang2017matterport3d} top-down views and evaluated on Path and PathArr settings only.
Full filtering criteria and real-world annotation pipeline details are in the Appendix.

\subsection{Multiview Counting}
\label{sec:multiview_counting}

Given several first-person frames of the same environment, the model must select the correct count of a queried object (\eg, ``How many chairs are in this area?''). Since no single view reveals the full layout, and the same object often appears across multiple frames, the model must construct a unified spatial representation that resolves both occlusions and cross-view duplicates.

\noindent\textbf{Imaginative perception target.} A top-down bird's-eye view (BEV) map aggregating all input views, making de-duplication explicit by mapping each object to a single spatial location. Ground-truth BEV maps are rendered from an overhead camera in the 3D scene.

\noindent\textbf{Data.}
Synthetic examples are generated via multi-camera and rotation trajectory types.
Real-world data is sourced from MessyTable~\cite{cai2020messytable} (fixed multi-camera rig; overhead image as ground-truth BEV) and ScanNet++~\cite{yeshwanth2023scannet++} (point-cloud BEV maps converted to photorealistic overhead images via Qwen Edit~\cite{wu2025qwenimagetechnicalreport}).
Questions are four-choice MCQ with distractors sampled near the true count.
The base training set contains $17{,}079$ synthetic examples; the mixed variant totals $19{,}499$.
We evaluate on a human-verified benchmark of $260$ samples.
Details on trajectory types, BEV rendering, and distractor sampling are in the Appendix.

\section{Method: Imaginative Perception Tokens}


The core of our approach is to enable Multimodal Language Models (MLLMs) to externalize spatial reasoning through \textbf{Imaginative Perception Tokens}.
Unlike standard textual chain-of-thought
or methods outsourcing visual imagination with an external visual generation model, our method requires the model to generate a visual representation of a \textit{non-observed} spatial configuration---such as an unseen viewpoint or an integrated top-down map---as a functional prerequisite for answering a spatial query.

\subsection{Problem Formalization}
Given an input context $\mathcal{C}$ consisting of one or more observed images $\mathcal{I}_{obs} = \{I_1, \dots, I_k\}$ and a spatial language query $Q$, the goal is to predict the correct answer $A$.
We decompose this into a two-stage generative process.
First, the model generates \textbf{imaginative perception tokens} $\hat{I}_{imag}$, representing the implied spatial structure requested by the task (e.g., the view from a new coordinate): $P(\hat{I}_{imag} | \mathcal{I}_{obs}, Q)$
Second, conditioned on this imaginative perception tokens $\hat{I}_{imag}$, the model produces the final answer: $P(A | \mathcal{I}_{obs}, Q, \hat{I}_{imag})$.

\subsection{Architecture}
We implement this approach using BAGEL~\cite{deng2025bagel}, a unified decoder-only transformer that natively supports interleaved multimodal understanding and generation.
BAGEL employs a {Mixture-of-Transformer-Experts (MoT)} design: 
the model utilizes two transformer experts, one optimized for multimodal understanding and another for generation.
Both operate on the same token sequence through shared self-attention at every layer.
Images are represented via two distinct paths. \textit{Understanding tokens} ($U$) are extracted via a SigLIP2~\cite{tschannen2025siglip} ViT encoder to capture semantic content, while \textit{Generation tokens} ($G$) are latent representations from a FLUX VAE used for high-fidelity synthesis.
Because all tokens (text, $U$, and $G$) coexist in a single shared context window, the model maintains lossless interaction between understanding and generation modules.

While BAGEL's standard generation tokens are typically used for open-ended text-to-image generation or editing, we repurpose this generative capacity for \textbf{spatial reasoning}. In our framework, the generation target is not a stylistic output but a precise \textbf{view imagination}---a visually grounded intermediate that represents the unobserved 3D structure of the scene.

\subsection{Training and Inference}

\noindent\textbf{Training Objective.}
We optimize the framework using a multi-task loss $\mathcal{L}_{total} = \lambda_{fm} \mathcal{L}_{fm} + \lambda_{lm} \mathcal{L}_{lm}$. The model is trained to jointly produce the imaginative perception and the final answer:
\begin{enumerate}
    \item \textbf{Flow-Matching Loss ($\mathcal{L}_{fm}$):} For the imaginative intermediate, BAGEL adopts the \textbf{Rectified Flow} method. The model learns to predict the velocity field $v_t$ required to transform Gaussian noise into the target latent $G_{gt}$ representing the unobserved view, conditioned on the preceding context $\mathcal{C}$:
    \begin{equation}
        \mathcal{L}_{fm} = \mathbb{E}_{t, G_0, \mathcal{C}} \left[ \| v_t(G_t | \mathcal{C}) - (G_{gt} - G_0) \|^2 \right]
    \end{equation}
    \item \textbf{Language Modeling Loss ($\mathcal{L}_{lm}$):} We minimize the negative log-likelihood of the final VQA answer tokens $A$, conditioned on the observed context and the ground-truth imaginative tokens:
    \begin{equation}
        \mathcal{L}_{lm} = - \sum_{i=1}^{|A|} \log P(a_i | \mathcal{C}, U_{gt}, G_{gt}, a_{<i})
    \end{equation}
\end{enumerate}

\noindent\textbf{Inference.} 
At inference time, the model operates in one of two modes depending on the task and configuration.
In the \textbf{text-only} mode, the model produces only a textual answer without generating any visual intermediate $A \sim P(A \mid \mathcal{C})$, serving as a baseline.
In the \textbf{imagination} mode, the model first performs iterative denoising over VAE tokens to produce the imaginative latent:
$\hat{G}_{imag} = \int_{0}^{1} v_t(G_t \mid \mathcal{C}) \, dt$
The decoded image $\hat{I}_{imag}$ is immediately re-encoded and appended to the context as both ViT understanding tokens and VAE generation tokens:
$\mathcal{C}' = \left[\mathcal{C},\, \text{ViT}(\hat{I}_{imag}),\, \text{VAE}(\hat{I}_{imag})\right]$
The model then attends to its own imagination to predict the final answer $A \sim P(A \mid \mathcal{C}')$.

\section{Experiments}

\begin{table}[t]
  \caption{\textbf{Main results.} Accuracy (\%) on AI2-THOR (in-domain) and different-environment (out-of-domain) benchmarks. PT reports the average across input settings (EgoDir/Path/PathArr for AI2-THOR; Real/Real+Arr for different environments). Text CoT generates a textual chain-of-thought before answering. IPT (Imaginative Perception Token) generates an intermediate image before answering. For our models, accuracy reports the maximum between answer-only and free-generation inference. Best per group in \textbf{bold}.}
  \vspace{-1em}
  \centering
  \setlength{\tabcolsep}{13pt}
  \begin{tabular}{l ccc cc}
    \toprule
    & \multicolumn{3}{c}{\textbf{AI2-THOR}} & \multicolumn{2}{c}{\textbf{Different Env.}} \\
    \cmidrule(lr){2-4} \cmidrule(lr){5-6}
    \textbf{Model}
      & PET & \footnotesize PT & MVC
      & \footnotesize PET & \footnotesize PT \\
    \midrule
    \multicolumn{6}{l}{\emph{VQA Models}} \\
    GPT-5            & \textbf{79.8} & \textbf{60.2} & 53.5 & \textbf{69.3} & \textbf{80.9} \\
    GPT-5.2          & 45.5 & 32.9 & 44.2 & 54.0 & 63.0 \\
    Gemini 2.5 Flash & 51.0 & 41.5 & 30.8 & 66.3 & 71.4 \\
    Gemini 3 Flash   & 55.0 & 42.3 & \textbf{56.9} & 51.3 & 83.2 \\
    InternVL3.5-8B   & 51.5 & 35.8 & 44.6 & 47.7 & 47.4 \\
    Qwen2.5-VL-7B   & 50.7 & 37.3 & 38.8 & 54.3 & 44.8 \\
    Qwen3-VL-8B     & 52.0 & 35.9 & 43.8 & 46.7 & 64.1 \\
    \midrule
    \multicolumn{6}{l}{\emph{Unified Models}} \\
    Janus-Pro-7B     & 51.8 & 33.5 & 33.1 & 44.7 & 35.3 \\
    Chameleon 7B     & 34.3 & 16.3 & 5.4 & 47.3 & 24.5 \\
    \midrule
    \multicolumn{6}{l}{\emph{Ours (fine-tuned BAGEL)}} \\
    Bagel (base)              & 40.3 & 29.9 & 35.4 & 62.7 & 42.7 \\
    Bagel (label-only)        & 97.5 & 65.7 & 63.9 & 82.0 & 54.7 \\
    + Text CoT                 & 83.1 & 49.7 & 62.3 & 70.3 & 52.2 \\
    + IPT                       & 96.8 & 49.0 & \textbf{67.3} & 87.0 & 57.5 \\
    + Mixed Training            & \textbf{97.8} & \textbf{66.7} & 62.3 & \textbf{87.7} & \textbf{58.6} \\
    \bottomrule
  \end{tabular}
  \label{tab:main_results}
\end{table}

We evaluate \emph{imaginative perception tokens} on the three spatial reasoning tasks introduced in \cref{sec:data}: \textbf{Perspective Taking (PET)}, \textbf{Path Tracing (PT)}, and \textbf{Multiview Counting (MVC)}. To enable controlled comparisons, we train all task-specific models on the AI2-THOR subset of each dataset. We additionally report transfer to cross-environment benchmarks (Habitat), real-world images, and external datasets. All tasks use multiple-choice evaluation with balanced answer distributions.

PT is evaluated under three input variants that provide increasing spatial cues: \textbf{EgoDir} (egocentric direction only), \textbf{Path} (top-down path overlay), and \textbf{PathArr} (path with directional arrows) and average accuracy reported. Unless otherwise stated, we report accuracy (\%) and use the same prompt formatting across baselines and our models.

\subsection{Setup}

\noindent\textbf{Baselines.}
We compare against two groups of models, evaluated zero-shot with task-specific prompts.
\emph{VQA models} include GPT-5, GPT-5.2, Gemini~2.5 Flash, Gemini~3 Flash, InternVL3.5-8B, Qwen2.5-VL-7B, and Qwen3-VL-8B.
\emph{Unified models} that support both understanding and generation include Janus-Pro-7B and Chameleon~7B.

\noindent\textbf{Our model variants.}
We fine-tune BAGEL~\cite{deng2025bagel} under several configurations to isolate the contribution of imagination supervision. Each fine-tuned model is task-specific and trained on a single task using AI2-THOR data only (unless noted otherwise):
\begin{itemize}
    \item \textbf{Bagel (base)}: pretrained model with no task-specific fine-tuning.
    \item \textbf{Bagel (label-only)}: fine-tuned with answer supervision only, with no intermediate thought.
    \item \textbf{+ Text CoT}: trained to generate a textual chain-of-thought describing the imagined spatial configuration before answering. Training CoTs are generated by GPT-5.1 using simulator ground-truth scene metadata.
    \item \textbf{+ IPT}: trained to generate an intermediate image (the imaginative perception token) before answering.
    \item \textbf{+ Mixed Training}: trained on a mixture of IPT examples (image-generation targets) and label-only examples (answer supervision only).
\end{itemize}

\noindent\textbf{Training details.}
We fine-tune BAGEL-7B-MoT with AdamW (lr $1{\times}10^{-5}$, $2{,}000$ warmup steps) on $8$ GPUs using FSDP bf16, following BAGEL~\cite{deng2025bagel} and ThinkMorph~\cite{gu2025thinkmorph}.
For multi-image inputs, each image is resized to $512{\times}512$.
Unless noted, IPTs use Latent-64 resolution.

\subsection{Main results}

Table~\ref{tab:main_results} reports results on our benchmarks.

\begin{takeawayblock}
    \textbf{Spatial reasoning remains difficult for current VLM and unified models.}
\end{takeawayblock}
Among the zero-shot baselines, GPT-5 is the strongest across nearly all settings, yet still trails our best fine-tuned variants on multiple in-distribution tasks.
Smaller open VLM models (InternVL3.5-8B, Qwen2.5-VL-7B, Qwen3-VL-8B) hover near chance on PET (50--52\%) and struggle on PT, indicating that these tasks are not solvable through superficial cues.
Unified models perform worse overall: Chameleon~7B drops to 34.3\% on PET and 5.4\% on MVC, suggesting that current unified designs often trade away understanding robustness in exchange for generation capability.

\begin{takeawayblock}
    \textbf{Answer supervision alone yields large gains and transfers across environments.}
\end{takeawayblock}
Bagel (label-only) substantially improves over Bagel (base) across all tasks, rising from 40.3\% to 97.5\% on AI2-THOR PET, from 29.9\% to 65.7\% on PT, and from 35.4\% to 63.9\% on MVC.
These improvements transfer: label-only reaches 82.0\% on Habitat PET, showing that spatial reasoning can be learned in simulation and generalized to new environments.

\begin{takeawayblock}
    \textbf{Imagination supervision helps most when language is a poor interface.}
\end{takeawayblock}
On MVC, IPT achieves the best accuracy (67.3\%), outperforming label-only (63.9\%) and Text CoT (62.3\%).
On different-environment PET (Habitat), IPT reaches 87.0\% (vs.\ 82.0\% for label-only), and Mixed Training improves further to 87.7\%.
On PT, Mixed Training achieves the best results on both synthetic (66.7\%) and real (58.6\%) benchmarks, outperforming label-only (65.7\% / 54.7\%) and all baselines.
IPT also improves real-world PT transfer (57.5\%) over label-only (54.7\%) and Text CoT (52.2\%).
Notably, IPT models are evaluated in \emph{answer-only} mode: the model does not generate an image at inference, yet the imagination targets during training strengthen internal spatial representations that transfer across environments.

\begin{takeawayblock}
    \textbf{Text CoT underperforms label-only and IPT.}
\end{takeawayblock}
Text CoT typically falls behind label-only (e.g., PET 83.1\% vs.\ 97.5\%, PT 49.7\% vs.\ 65.7\%) and also behind IPT (e.g., MVC 62.3\% vs.\ 67.3\%, PET 83.1\% vs.\ 96.8\%).
Compared to label-only, the Text CoT objective forces the model to allocate capacity to generating long spatial descriptions during fine-tuning, which competes with answer prediction.
Compared to IPT, the gap reflects a modality mismatch: viewpoint changes, occlusions, and cross-view correspondences are difficult to serialize into natural language, and the resulting textual traces introduce noise rather than useful structure. IPT represents these relationships directly in the visual modality where they are naturally expressed.

\subsection{Ablations}
\label{sec:exp_ablation}

\begin{figure}[t]
  \centering
  \includegraphics[width=\linewidth, height=0.5\linewidth]{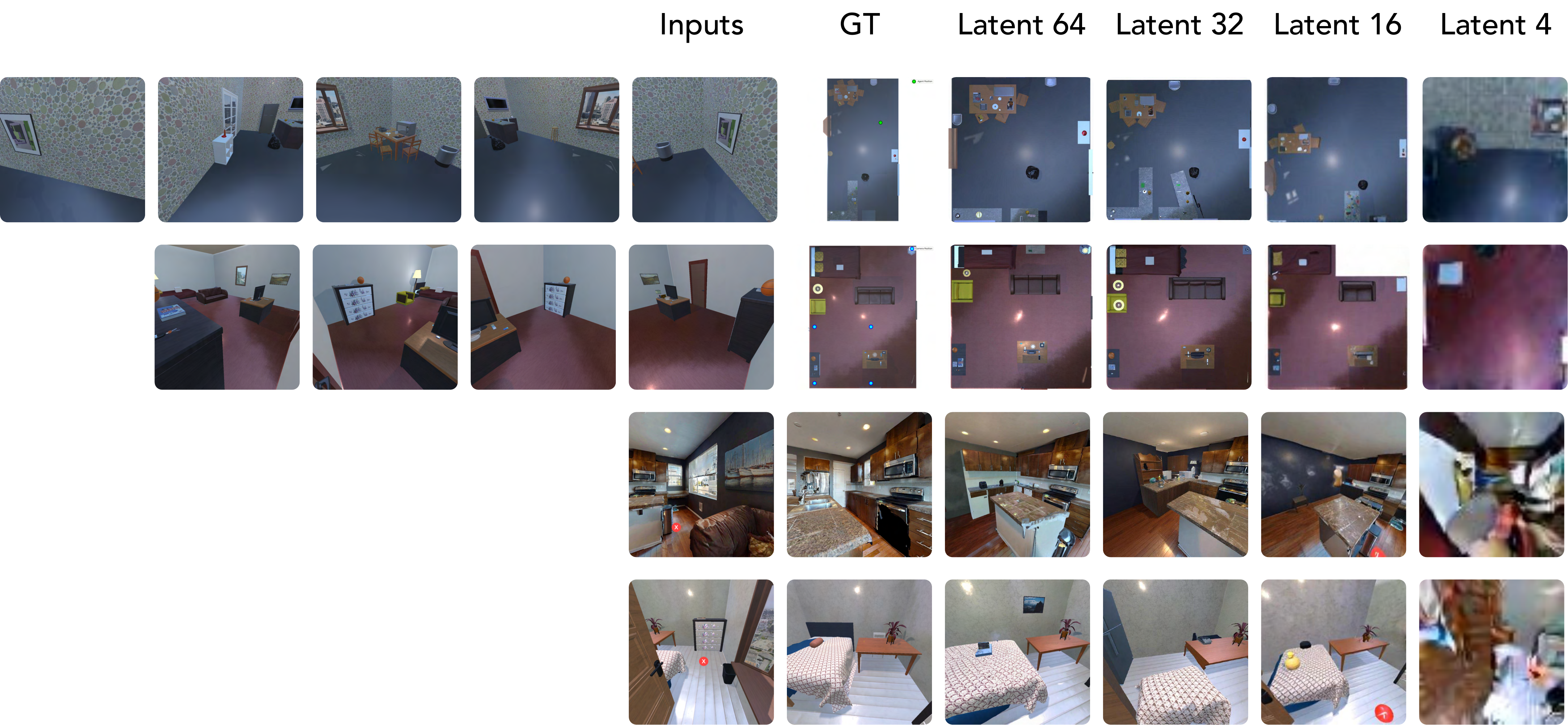}
  \caption{Qualitative examples of model-generated imaginative perception tokens. Top two rows: MVC example showing imagined top-down BEV maps. Bottom: PET examples showing imagined novel viewpoints. From left to right, imagination resolution increases from Latent-4 ($64\times 64$) to Latent-64 ($1024\times 1024$). Higher resolution produces sharper and more spatially faithful imaginations, preserving object identities and relative positions needed for downstream reasoning.}
  \label{fig:ablation_latent}
\end{figure}
\begin{table}[t]
  \caption{\textbf{Ablation on latent size.} Accuracy (\%) with w/ Thought inference mode at different imagination resolutions. Best per column in \textbf{bold}.}
  \vspace{-1em}
  \centering
  \setlength{\tabcolsep}{10pt}
  \small
  \begin{tabular}{l c cc c}
    \toprule
    & & \multicolumn{2}{c}{\textbf{PET}} & \textbf{MVC} \\
    \cmidrule(lr){3-4} \cmidrule(lr){5-5}
    \textbf{Latent Size} & \textbf{Resolution}
      & \footnotesize AI2-THOR & \footnotesize Habitat
      & \footnotesize AI2-THOR \\
    \midrule
    Latent-4  & $64\times64$     & 87.4 & 73.3 & 53.5 \\
    Latent-16 & $256\times256$   & 95.3 & 81.0 & 56.2 \\
    Latent-32 & $512\times512$   & 95.0 & \textbf{87.0} & 58.9 \\
    Latent-64 & $1024\times1024$ & \textbf{96.8} & 83.3 & \textbf{63.1} \\
    \bottomrule
  \end{tabular}
  \label{tab:ablation_latent}
\end{table}

\begin{takeawayblock}
    \textbf{Latent resolution controls imagination quality and downstream accuracy.}
\end{takeawayblock}
\Cref{tab:ablation_latent,fig:ablation_latent} ablate IPT resolution on PET and MVC.
At Latent-4 ($64\times64$), imaginations are blurry and lose spatial detail; at Latent-64 ($1024\times1024$), imaginations become sharper and more spatially faithful, preserving object identities and relative positions.
Quantitatively, increasing resolution from Latent-4 to Latent-64 improves AI2-THOR PET from 87.4\% to 96.8\% and MVC from 53.5\% to 63.1\%.
Habitat PET peaks at Latent-32 (87.0\%) and drops slightly at Latent-64 (83.3\%), suggesting mild overfitting to AI2-THOR appearance statistics at the highest resolution.

\begin{table}[t]
  \caption{\textbf{Ablation on thought modality and inference mode.} Accuracy (\%) on AI2-THOR benchmarks (PT uses EgoDir variant). We compare Text CoT vs.\ IPT training and vary inference mode: generate thought then answer (w/ text/image), answer directly (answer-only), or condition on ground-truth (w/ GT image).}
  \vspace{-1em}
  \centering
  \setlength{\tabcolsep}{14pt}
  \small
  \begin{tabular}{ll ccc}
    \toprule
    \textbf{Training} & \textbf{Inference} & \textbf{PET} & \textbf{PT} & \textbf{MVC} \\
    \midrule
    Text CoT   & w/ text      & 83.1 & 53.1 & 61.5 \\
    Text CoT   & answer-only  & 78.3 & 55.8 & 62.3 \\
    \midrule
    IPT        & w/ image     & 96.8 & 50.4 & \textbf{67.3} \\
    IPT        & answer-only  & 96.8 & 61.1 & 62.3 \\
    \midrule
    IPT        & w/ GT image  & 96.7 & \textbf{86.7} & \textbf{67.3} \\
    \bottomrule
  \end{tabular}
  \label{tab:ablation_thought}
\end{table}
\vspace{-1em}
\paragraph{Thought modality and inference mode.}
Table~\ref{tab:ablation_thought} ablates the training signal (Text CoT vs.\ IPT) and inference mode (generate thought, answer-only, or oracle GT).

\begin{takeawayblock}
    \textbf{IPT training builds stronger spatial representations than Text CoT.}
\end{takeawayblock}
On PT, IPT with answer-only inference (61.1\%) outperforms Text CoT with answer-only inference (55.8\%) by 5.3 points.
On MVC, IPT with image generation (63.1\%) outperforms Text CoT with text generation (61.5\%).

\begin{takeawayblock}
    \textbf{Imagination supervision is useful, but explicit generation is not required at inference.}
\end{takeawayblock}
For IPT models, answer-only mostly outperforms generating the imagination explicitly: on PT, answer-only reaches 61.1\% vs.\ 50.4\% with generation.
For Text CoT, generating the chain-of-thought also slightly underperforms answer-only (53.1\% vs.\ 55.8\% on PT), though the gap is smaller than for IPT.
This asymmetry suggests that producing faithful imaginations is harder than producing text descriptions, and imperfect generations can mislead downstream reasoning. However, training with imagination targets remains valuable: answer-only IPT matches GPT-5 on PT (61.1\%).

\begin{takeawayblock}
    \textbf{Ground-truth imaginations reveal headroom.}
\end{takeawayblock}
When given ground-truth imaginations instead of model-generated ones, PT accuracy jumps from 50.4\% to 86.7\% (+36.3) and MVC rises from 63.1\% to 67.3\% (+4.2).
The large PT gap indicates that imagination quality is the dominant bottleneck for path tracing; for PET, model-generated imaginations nearly match GT (96.8\% vs.\ 96.7\%), leaving little room for improvement.

\begin{table}[t]
  \caption{\textbf{Transfer to similar external benchmarks.} Accuracy (\%). SAT tests perspective taking and MessyTable tests multiview counting, both in domains unseen during training. Best in \textbf{bold}.}
  \vspace{-1em}
  \centering
  \setlength{\tabcolsep}{19pt}
  \small
  \begin{tabular}{l cc}
    \toprule
    & \textbf{PET} & \textbf{MVC} \\
    \cmidrule(lr){2-2} \cmidrule(lr){3-3}
    \textbf{Model}
      & \footnotesize SAT (66)
      & \footnotesize MessyTable (200) \\
    \midrule
    Bagel (base)       & 34.9 & 29.0 \\
    Bagel (label-only) & 59.1 & 32.5 \\
    + Text CoT         & 50.0 & 30.0 \\
    + IPT              & 57.6 & 28.5 \\
    + Mixed Training   & \textbf{63.6} & \textbf{37.0} \\
    \bottomrule
  \end{tabular}
  \label{tab:transfer}
\end{table}

\begin{takeawayblock}
    \textbf{IPT transfers to aligned external benchmarks.}
\end{takeawayblock}
Table~\ref{tab:transfer} evaluates transfer to external benchmarks that test similar spatial capabilities: SAT~\cite{ray2025satdynamicspatialaptitude} (perspective-taking subset) and MessyTable~\cite{cai2020messytable} (multiview counting).
On SAT, Bagel (label-only) improves from 34.9\% to 59.1\% over Bagel (base), and Mixed Training further improves to 63.6\%.
On MessyTable, Mixed Training reaches 37.0\%, up from 29.0\% for Bagel (base).

\begin{table}[t]
  \caption{\textbf{Does our data help on other spatial tasks?} Accuracy (\%) on benchmarks beyond our training task categories. Fine-tuning on our AI2-THOR MVC data consistently improves over Bagel (base), suggesting that the spatial reasoning learned from our datasets transfers broadly. Best per column in \textbf{bold}.}
  \vspace{-1em}
  \centering
  \setlength{\tabcolsep}{8pt}
  \small
  \begin{tabular}{l ccc}
    \toprule
    \textbf{Model}
      & \footnotesize ScanNet (200)
      & \footnotesize MindCube (200)
      & \footnotesize All-Angles (170) \\
    \midrule
    \multicolumn{4}{l}{\emph{VQA Models}} \\
    GPT-5            & 58.5 & \textbf{67.3} & \textbf{67.9} \\
    GPT-5.2          & 48.5 & 37.5 & 29.4 \\
    Gemini 2.5 Flash & 48.0 & 50.3 & 37.9 \\
    Gemini 3 Flash   & 62.5 & 56.5 & 64.2 \\
    InternVL3.5-8B   & 53.5 & 42.1 & 54.8 \\
    Qwen2.5-VL-7B   & \textbf{63.5} & 47.8 & 51.8 \\
    Qwen3-VL-8B     & 62.5 & 34.5 & 42.3 \\
    \midrule
    \multicolumn{4}{l}{\emph{Unified Models}} \\
    Janus-Pro-7B     & 39.5 & 42.0 & 45.0 \\
    Chameleon 7B     & 5.5 & 25.4 & 17.7 \\
    \midrule
    \multicolumn{4}{l}{\emph{Ours (fine-tuned BAGEL)}} \\
    Bagel (base)       & 40.5 & 39.5 & 40.0 \\
    Bagel (fine-tuned) & \textbf{52.0} & \textbf{47.5} & \textbf{50.0} \\
    \bottomrule
  \end{tabular}
  \label{tab:ablation_data}
\end{table}

\begin{takeawayblock}
    \textbf{Training with our data improves performance on other spatial benchmarks.}
\end{takeawayblock}
Finally, we test whether our training data improves spatial reasoning on tasks with different structures: ScanNet~\cite{dai2017scannet} (in-the-wild multiview counting), MindCube~\cite{yin2025mindcube} (abstract geometric reasoning), and All-Angles-Bench~\cite{yeh2025seeing} (cross-view matching on EgoHumans~\cite{khirodkar2023egohumans}).
Because IPTs are task-specific by construction (e.g., rotated views for PET, bird's-eye paths for PT), they do not directly transfer to these settings. We therefore fine-tune on AI2-THOR MVC using answer supervision only.
Bagel (fine-tuned) consistently improves over Bagel (base) across all three benchmarks (40.5\%$\to$52.0\% on ScanNet, 39.5\%$\to$47.5\% on MindCube, 40.0\%$\to$50.0\% on All-Angles), indicating that our simulator data builds broadly useful spatial representations even when the specific imaginative token target changes.

\section{Conclusion}
\label{sec:conclusion}

We introduced Imaginative Perception Tokens, intermediate visual representations that externalize spatial reasoning about unobserved structure in multimodal language models.
We designed three tasks, Perspective Taking, Path Tracing, and Multiview Counting, that specifically require constructing missing spatial information, and built datasets with ground-truth intermediate imaginations for each.
Experiments on a unified MLLM backbone show that training with imagination supervision consistently improves spatial reasoning over both label-only and text chain-of-thought baselines, even when the imagination is not explicitly generated at inference.
Ablations further reveal that imagination quality directly governs downstream accuracy, with ground-truth intermediates yielding large headroom over current generation quality.

\section*{Acknowledgements}
This work was partially supported by Samsung and CoCoSys. We thank Zhe Zhou for helping clean and filter our datasets and benchmarks.
\clearpage

%
%
\bibliographystyle{splncs04}
\bibliography{main}

@String(CVPR  = {IEEE Conf. Comput. Vis. Pattern Recog.})

@String(ICCV  = {Int. Conf. Comput. Vis.})

@String(NeurIPS = {Adv. Neural Inform. Process. Syst.})

@String(ICLR  = {Int. Conf. Learn. Represent.})

@String(CVPR  = {CVPR})

@String(ICCV  = {ICCV})

@String(NeurIPS = {NeurIPS})

@String(ICLR  = {ICLR})

@misc{khirodkar2023egohumans,
      title={EgoHumans: An Egocentric 3D Multi-Human Benchmark},
      author={Rawal Khirodkar and Aayush Bansal and Lingni Ma and Richard Newcombe and Minh Vo and Kris Kitani},
      year={2023},
      eprint={2305.16487},
      archivePrefix={arXiv},
      primaryClass={cs.CV},
}

@inproceedings{yang2019spatialsense,
  author    = {Yang, Kaiyu and Russakovsky, Olga and Deng, Jia},
  title     = {{SpatialSense}: An Adversarially Crowdsourced Benchmark for Spatial Relation Recognition},
  booktitle = ICCV,
  year      = {2019}
}

@article{liu2022vsr,
  author  = {Liu, Fangyu and Emerson, Guy and Collier, Nigel},
  title   = {Visual Spatial Reasoning},
  journal = {arXiv preprint arXiv:2205.00363},
  year    = {2022}
}

@inproceedings{kamath2023whatsup,
  author    = {Kamath, Amita and Hessel, Jack and Chang, Kai-Wei},
  title     = {What's ``up'' with Vision-Language Models? {I}nvestigating their Struggle with Spatial Reasoning},
  booktitle = {Proceedings of the Conference on Empirical Methods in Natural Language Processing (EMNLP)},
  year      = {2023}
}

@inproceedings{ma20253dsrbench,
  author    = {Ma, Wufei and Chen, Haoyu and Zhang, Guofeng and Chou, Yu-Cheng and Chen, Jieneng and de Melo, Celso M. and Yuille, Alan},
  title     = {{3DSRBench}: A Comprehensive {3D} Spatial Reasoning Benchmark},
  booktitle = ICCV,
  year      = {2025}
}

@article{wei2022cot,
  author  = {Wei, Jason and Wang, Xuezhi and Schuurmans, Dale and Bosma, Maarten and Ichter, Brian and Xia, Fei and Chi, Ed and Le, Quoc and Zhou, Denny},
  title   = {Chain-of-Thought Prompting Elicits Reasoning in Large Language Models},
  journal = {arXiv preprint arXiv:2201.11903},
  year    = {2022}
}

@article{gu2025thinkmorph,
  author  = {Gu, Jiawei and Hao, Yunzhuo and Wang, Huichen Will and Li, Linjie and Shieh, Michael Qizhe and Choi, Yejin and Krishna, Ranjay and Cheng, Yu},
  title   = {{ThinkMorph}: Emergent Properties in Multimodal Interleaved Chain-of-Thought Reasoning},
  journal = {arXiv preprint arXiv:2510.27492},
  year    = {2025}
}

@misc{openai2025thinkingimages,
  author       = {{OpenAI}},
  title        = {Thinking with Images},
  howpublished = {OpenAI Blog},
  year         = {2025},
  url          = {https://openai.com/index/thinking-with-images/}
}

@article{hu2024visualsketchpad,
  author  = {Hu, Yushi and Shi, Weijia and Fu, Xingyu and Roth, Dan and Ostendorf, Mari and Zettlemoyer, Luke and Smith, Noah A. and Krishna, Ranjay},
  title   = {Visual Sketchpad: Sketching as a Visual Chain of Thought for Multimodal Language Models},
  journal = {arXiv preprint arXiv:2406.09403},
  year    = {2024}
}

@article{li2025mvot,
  author  = {Li, Chengzu and Wu, Wenshan and Zhang, Huanyu and Xia, Yan and Mao, Shaoguang and Dong, Li and Vuli{\'c}, Ivan and Wei, Furu},
  title   = {Imagine while Reasoning in Space: Multimodal Visualization-of-Thought},
  journal = {arXiv preprint arXiv:2501.07542},
  year    = {2025}
}

@article{bigverdi2024aurora,
  author  = {Bigverdi, Mahtab and Luo, Zelun and Hsieh, Cheng-Yu and Shen, Ethan and Chen, Dongping and Shapiro, Linda G. and Krishna, Ranjay},
  title   = {Perception Tokens Enhance Visual Reasoning in Multimodal Language Models},
  journal = {arXiv preprint arXiv:2412.03548},
  year    = {2024}
}

@article{yang2025mirage,
  author  = {Yang, Zeyuan and Yu, Xueyang and Chen, Delin and Shen, Maohao and Gan, Chuang},
  title   = {Machine Mental Imagery: Empower Multimodal Reasoning with Latent Visual Tokens},
  journal = {arXiv preprint arXiv:2506.17218},
  year    = {2025}
}

@article{ray2025mulltokens,
  author  = {Ray, Arijit and Abdelkader, Ahmed and Mao, Chengzhi and Plummer, Bryan A. and Saenko, Kate and Krishna, Ranjay and Guibas, Leonidas and Chu, Wen-Sheng},
  title   = {Mull-Tokens: Modality-Agnostic Latent Thinking},
  journal = {arXiv preprint arXiv:2512.10941},
  year    = {2025}
}

@article{tschannen2025siglip,
  title={Siglip 2: Multilingual vision-language encoders with improved semantic understanding, localization, and dense features},
  author={Tschannen, Michael and Gritsenko, Alexey and Wang, Xiao and Naeem, Muhammad Ferjad and Alabdulmohsin, Ibrahim and Parthasarathy, Nikhil and Evans, Talfan and Beyer, Lucas and Xia, Ye and Mustafa, Basil and others},
  journal={arXiv preprint arXiv:2502.14786},
  year={2025}
}

@article{yang2025visual,
  title={Visual spatial tuning},
  author={Yang, Rui and Zhu, Ziyu and Li, Yanwei and Huang, Jingjia and Yan, Shen and Zhou, Siyuan and Liu, Zhe and Li, Xiangtai and Li, Shuangye and Wang, Wenqian and others},
  journal={arXiv preprint arXiv:2511.05491},
  year={2025}
}

@article{kolve2017ai2,
  title={Ai2-thor: An interactive 3d environment for visual ai},
  author={Kolve, Eric and Mottaghi, Roozbeh and Han, Winson and VanderBilt, Eli and Weihs, Luca and Herrasti, Alvaro and Deitke, Matt and Ehsani, Kiana and Gordon, Daniel and Zhu, Yuke and others},
  journal={arXiv preprint arXiv:1712.05474},
  year={2017}
}

@misc{puig2023habitat3,
  title  = {Habitat 3.0: A Co-Habitat for Humans, Avatars and Robots},
  author = {Xavi Puig and Eric Undersander and Andrew Szot and Mikael Dallaire Cote and Ruslan Partsey and Jimmy Yang and Ruta Desai and Alexander William Clegg and Michal Hlavac and Tiffany Min and Theo Gervet and Vladimír Vondruš and Vincent-Pierre Berges and John Turner and Oleksandr Maksymets and Zsolt Kira and Mrinal Kalakrishnan and Jitendra Malik and Devendra Singh Chaplot and Unnat Jain and Dhruv Batra and Akshara Rai and Roozbeh Mottaghi},
  year={2023},
  archivePrefix={arXiv},
}

@inproceedings{szot2021habitat,
  title     =     {Habitat 2.0: Training Home Assistants to Rearrange their Habitat},
  author    =     {Andrew Szot and Alex Clegg and Eric Undersander and Erik Wijmans and Yili Zhao and John Turner and Noah Maestre and Mustafa Mukadam and Devendra Chaplot and Oleksandr Maksymets and Aaron Gokaslan and Vladimir Vondrus and Sameer Dharur and Franziska Meier and Wojciech Galuba and Angel Chang and Zsolt Kira and Vladlen Koltun and Jitendra Malik and Manolis Savva and Dhruv Batra},
  booktitle =     {Advances in Neural Information Processing Systems (NeurIPS)},
  year      =     {2021}
}

@article{yeh2025seeing,
  title={Seeing from another perspective: Evaluating multi-view understanding in mllms},
  author={Yeh, Chun-Hsiao and Wang, Chenyu and Tong, Shengbang and Cheng, Ta-Ying and Wang, Ruoyu and Chu, Tianzhe and Zhai, Yuexiang and Chen, Yubei and Gao, Shenghua and Ma, Yi},
  journal={arXiv preprint arXiv:2504.15280},
  year={2025}
}

@inproceedings{cai2020messytable,
  title={Messytable: Instance association in multiple camera views},
  author={Cai, Zhongang and Zhang, Junzhe and Ren, Daxuan and Yu, Cunjun and Zhao, Haiyu and Yi, Shuai and Yeo, Chai Kiat and Change Loy, Chen},
  booktitle={European Conference on Computer Vision},
  pages={1--16},
  year={2020},
  organization={Springer}
}

@misc{ray2025satdynamicspatialaptitude,
      title={SAT: Dynamic Spatial Aptitude Training for Multimodal Language Models}, 
      author={Arijit Ray and Jiafei Duan and Ellis Brown and Reuben Tan and Dina Bashkirova and Rose Hendrix and Kiana Ehsani and Aniruddha Kembhavi and Bryan A. Plummer and Ranjay Krishna and Kuo-Hao Zeng and Kate Saenko},
      year={2025},
      eprint={2412.07755},
      archivePrefix={arXiv},
      primaryClass={cs.CV},
      url={https://arxiv.org/abs/2412.07755}, 
    }

@inproceedings{habitat19iccv,
  title     =     {Habitat: {A} {P}latform for {E}mbodied {AI} {R}esearch},
  author    =     {{Manolis Savva*} and {Abhishek Kadian*} and {Oleksandr Maksymets*} and Yili Zhao and Erik Wijmans and Bhavana Jain and Julian Straub and Jia Liu and Vladlen Koltun and Jitendra Malik and Devi Parikh and Dhruv Batra},
  booktitle =     {Proceedings of the IEEE/CVF International Conference on Computer Vision (ICCV)},
  year      =     {2019}
}

@article{deng2025bagel,
  author  = {Deng, Chaorui and Zhu, Deyao and Li, Kunchang and Gou, Chenhui and Li, Feng and Wang, Zeyu and Zhong, Shu and Yu, Weihao and Nie, Xiaonan and Song, Ziang and Shi, Guang and Fan, Haoqi},
  title   = {Emerging Properties in Unified Multimodal Pretraining},
  journal = {arXiv preprint arXiv:2505.14683},
  year    = {2025}
}

@article{chameleon2024chameleon,
  author  = {{Chameleon Team}},
  title   = {Chameleon: Mixed-Modal Early-Fusion Foundation Models},
  journal = {arXiv preprint arXiv:2405.09818},
  year    = {2024}
}

@inproceedings{esser2021taming,
  title={Taming transformers for high-resolution image synthesis},
  author={Esser, Patrick and Rombach, Robin and Ommer, Bjorn},
  booktitle={Proceedings of the IEEE/CVF conference on computer vision and pattern recognition},
  pages={12873--12883},
  year={2021}
}

@inproceedings{yang2024depth,
  title={Depth anything: Unleashing the power of large-scale unlabeled data},
  author={Yang, Lihe and Kang, Bingyi and Huang, Zilong and Xu, Xiaogang and Feng, Jiashi and Zhao, Hengshuang},
  booktitle={Proceedings of the IEEE/CVF conference on computer vision and pattern recognition},
  pages={10371--10381},
  year={2024}
}

@inproceedings{wang2025site,
  title={Site: towards spatial intelligence thorough evaluation},
  author={Wang, Wenqi and Tan, Reuben and Zhu, Pengyue and Yang, Jianwei and Yang, Zhengyuan and Wang, Lijuan and Kolobov, Andrey and Gao, Jianfeng and Gong, Boqing},
  booktitle={Proceedings of the IEEE/CVF International Conference on Computer Vision},
  pages={9058--9069},
  year={2025}
}

@article{jia2025omnispatial,
  title={Omnispatial: Towards comprehensive spatial reasoning benchmark for vision language models},
  author={Jia, Mengdi and Qi, Zekun and Zhang, Shaochen and Zhang, Wenyao and Yu, Xinqiang and He, Jiawei and Wang, He and Yi, Li},
  journal={arXiv preprint arXiv:2506.03135},
  year={2025}
}

@article{wu2025spatialscore,
  title={Spatialscore: Towards unified evaluation for multimodal spatial understanding},
  author={Wu, Haoning and Huang, Xiao and Chen, Yaohui and Zhang, Ya and Wang, Yanfeng and Xie, Weidi},
  journal={arXiv e-prints},
  pages={arXiv--2505},
  year={2025}
}

@article{wang2025spatialviz,
  title={Spatialviz-bench: An mllm benchmark for spatial visualization},
  author={Wang, Siting and Pei, Minnan and Sun, Luoyang and Deng, Cheng and Shao, Kun and Tian, Zheng and Zhang, Haifeng and Wang, Jun},
  journal={arXiv preprint arXiv:2507.07610},
  year={2025}
}

@article{xie2025showo2,
  author  = {Xie, Jinheng and Yang, Zhenheng and Shou, Mike Zheng},
  title   = {Show-o2: Improved Native Unified Multimodal Models},
  journal = {arXiv preprint arXiv:2506.15564},
  year    = {2025}
}

@article{chen2025januspro,
  author  = {Chen, Xiaokang and Wu, Zhiyu and Liu, Xingchao and Pan, Zizheng and Liu, Wen and Xie, Zhenda and Yu, Xingkai and Ruan, Chong},
  title   = {Janus-Pro: Unified Multimodal Understanding and Generation with Data and Model Scaling},
  journal = {arXiv preprint arXiv:2501.17811},
  year    = {2025}
}

@article{li2025viewspatialbench,
  author  = {Li, Linnan and Chen, Xiaoyu and Chen, Peng and others},
  title   = {{ViewSpatial-Bench}: Evaluating Multi-Perspective Spatial Understanding of Vision-Language Models},
  journal = {arXiv preprint arXiv:2505.21500},
  year    = {2025}
}

@article{yang2025thinkinginspace,
  author  = {Yang, Jihan and Yang, Shusheng and Gupta, Anjali W. and Han, Rilyn and Fei{-}Fei, Li and Xie, Saining},
  title   = {Thinking in Space: How Multimodal Large Language Models See, Remember, and Recall Spaces},
  journal = {arXiv preprint arXiv:2412.14171},
  year    = {2025}
}

@article{yang2025mmsibench,
  author  = {Yang, Sihan and Xu, Runsen and Xie, Yiman and others},
  title   = {{MMSI-Bench}: A Benchmark for Multi-Image Spatial Intelligence},
  journal = {arXiv preprint arXiv:2505.23764},
  year    = {2025}
}

@article{yin2025mindcube,
  author  = {Yin, Baiqiao and Wang, Qineng and Zhang, Pingyue and others},
  title   = {Spatial Mental Modeling from Limited Views},
  journal = {arXiv preprint arXiv:2506.21458},
  year    = {2025}
}

@inproceedings{dumery2025countingstackedobjects,
  author    = {Dumery, Corentin and Ett{\'e}, Noa and Fan, Aoxiang and Li, Ren and Xu, Jingyi and Le, Hieu and Fua, Pascal},
  title     = {Counting Stacked Objects},
  booktitle = ICCV,
  year      = {2025}
}

@article{brown2025trainontestset,
  author    = {Brown, Ellis and Yang, Jihan and Yang, Shusheng and Fergus, Rob and Xie, Saining},
  title     = {Benchmark Designers Should ``Train on the Test Set'' to Expose Exploitable Non-Visual Shortcuts},
  journal   = {arXiv preprint arXiv:2511.04655},
  year      = {2025}
}

@inproceedings{deitke2025molmo,
  title     = {Molmo and PixMo: Open Weights and Open Data for State-of-the-Art Vision-Language Models},
  author    = {Deitke, Matt and Clark, Christopher and Lee, Sangho and Tripathi, Rohun and Yang, Yue and Park, Jae Sung and Salehi, Mohammadreza and Muennighoff, Niklas and Lo, Kyle and Soldaini, Luca and Lu, Jiasen and Anderson, Taira and Bransom, Erin and Ehsani, Kiana and Ngo, Huong and Chen, YenSung and Patel, Ajay and Yatskar, Mark and Callison-Burch, Chris and Head, Andrew and Hendrix, Rose and Bastani, Favyen and VanderBilt, Eli and Lambert, Nathan and Chou, Yvonne and Chheda, Arnavi and Sparks, Jenna and Skjonsberg, Sam and Schmitz, Michael and Sarnat, Aaron and Bischoff, Byron and Walsh, Pete and Newell, Chris and Wolters, Piper and Gupta, Tanmay and Zeng, Kuo-Hao and Borchardt, Jon and Groeneveld, Dirk and Nam, Crystal and Lebrecht, Sophie and Wittlif, Caitlin and Schoenick, Carissa and Michel, Oscar and Krishna, Ranjay and Weihs, Luca and Smith, Noah A. and Hajishirzi, Hannaneh and Girshick, Ross and Farhadi, Ali and Kembhavi, Aniruddha},
  booktitle = CVPR,
  year      = {2025},
  url       = {https://arxiv.org/abs/2409.17146}
}

@article{clark2026molmo2,
  title   = {Molmo2: Open Weights and Data for Vision-Language Models with Video Understanding and Grounding},
  author  = {Clark, Christopher and Zhang, Jieyu and Ma, Zixian and Park, Jae Sung and Salehi, Mohammadreza and Tripathi, Rohun and Lee, Sangho and Ren, Zhongzheng and Kim, Chris Dongjoo and Yang, Yinuo and Shao, Vincent and Yang, Yue and Huang, Weikai and Gao, Ziqi and Anderson, Taira and Zhang, Jianrui and Jain, Jitesh and Stoica, George and Han, Winson and Farhadi, Ali and Krishna, Ranjay},
  journal = {arXiv preprint arXiv:2601.10611},
  year    = {2026},
  url     = {https://arxiv.org/abs/2601.10611}
}

@article{bai2025qwen3vl,
  title   = {Qwen3-VL Technical Report},
  author  = {Bai, Shuai and Cai, Yuxuan and Chen, Ruizhe and Chen, Keqin and Chen, Xionghui and Cheng, Zesen and Deng, Lianghao and Ding, Wei and Gao, Chang and Ge, Chunjiang and Ge, Wenbin and Guo, Zhifang and Huang, Qidong and Huang, Jie and Huang, Fei and Hui, Binyuan and Jiang, Shutong and Li, Zhaohai and Li, Mingsheng and Li, Mei and Li, Kaixin and Lin, Zicheng and Lin, Junyang and Liu, Xuejing and Liu, Jiawei and Liu, Chenglong and Liu, Yang and Liu, Dayiheng and Liu, Shixuan and Lu, Dunjie and Luo, Ruilin and Lv, Chenxu and Men, Rui and Meng, Lingchen and Ren, Xuancheng and Ren, Xingzhang and Song, Sibo and Sun, Yuchong and Tang, Jun and Tu, Jianhong and Wan, Jianqiang and Wang, Peng and Wang, Pengfei and Wang, Qiuyue and Wang, Yuxuan and Xie, Tianbao and Xu, Yiheng and Xu, Haiyang and Xu, Jin and Yang, Zhibo and Yang, Mingkun and Yang, Jianxin and Yang, An and Yu, Bowen and Zhang, Fei and Zhang, Hang and Zhang, Xi and Zheng, Bo and Zhong, Humen and Zhou, Jingren and Zhou, Fan and Zhou, Jing and Zhu, Yuanzhi and Zhu, Ke},
  journal = {arXiv preprint arXiv:2511.21631},
  year    = {2025},
  url     = {https://arxiv.org/abs/2511.21631}
}

@misc{hu2023tifa,
  title={TIFA: Accurate and Interpretable Text-to-Image Faithfulness Evaluation with Question Answering},
  author={Yushi Hu and Benlin Liu and Jungo Kasai and Yizhong Wang and Mari Ostendorf and Ranjay Krishna and Noah A Smith},
  year={2023},
  eprint={2303.11897},
  archivePrefix={arXiv},
  primaryClass={cs.CV},
  url={https://arxiv.org/abs/2303.11897}
}

@misc{chang2017matterport3d,
  title={Matterport3D: Learning from RGB-D Data in Indoor Environments},
  author={Angel Chang and Angela Dai and Thomas Funkhouser and Maciej Halber and Matthias Nie{\ss}ner and Manolis Savva and Shuran Song and Andy Zeng and Yinda Zhang},
  year={2017},
  eprint={1709.06158},
  archivePrefix={arXiv},
  primaryClass={cs.CV},
  url={https://arxiv.org/abs/1709.06158}
}

@misc{dai2017scannet,
  title={ScanNet: Richly-annotated 3D Reconstructions of Indoor Scenes},
  author={Angela Dai and Angel X. Chang and Manolis Savva and Maciej Halber and Thomas Funkhouser and Matthias Nie{\ss}ner},
  year={2017},
  eprint={1702.04405},
  archivePrefix={arXiv},
  primaryClass={cs.CV},
  url={https://arxiv.org/abs/1702.04405}
}

@misc{deitke2022procthor,
  title={ProcTHOR: Large-Scale Embodied AI Using Procedural Generation},
  author={Matt Deitke and Eli VanderBilt and Alvaro Herrasti and Luca Weihs and Jordi Salvador and Kiana Ehsani and Winson Han and Eric Kolve and Ali Farhadi and Aniruddha Kembhavi and Roozbeh Mottaghi},
  year={2022},
  eprint={2206.06994},
  archivePrefix={arXiv},
  primaryClass={cs.AI},
  url={https://arxiv.org/abs/2206.06994}
}

@inproceedings{zhang2026theoryofspace,
  title     = {Theory of Space: Can Foundation Models Construct Spatial Beliefs through Active Exploration?},
  author    = {Zhang, Pingyue and Huang, Zihan and Wang, Yue and Zhang, Jieyu and Xue, Letian and Wang, Zihan and Wang, Qineng and Chandrasegaran, Keshigeyan and Zhang, Ruohan and Choi, Yejin and Krishna, Ranjay and Wu, Jiajun and Fei-Fei, Li and Li, Manling},
  booktitle = {International Conference on Learning Representations (ICLR)},
  year      = {2026},
}

@inproceedings{yeshwanth2023scannet++,
  title={Scannet++: A high-fidelity dataset of 3d indoor scenes},
  author={Yeshwanth, Chandan and Liu, Yueh-Cheng and Nie{\ss}ner, Matthias and Dai, Angela},
  booktitle={Proceedings of the IEEE/CVF International Conference on Computer Vision},
  pages={12--22},
  year={2023}
}

@misc{wu2025qwenimagetechnicalreport,
      title={Qwen-Image Technical Report}, 
      author={Chenfei Wu and Jiahao Li and Jingren Zhou and Junyang Lin and Kaiyuan Gao and Kun Yan and Sheng-ming Yin and Shuai Bai and Xiao Xu and Yilei Chen and Yuxiang Chen and Zecheng Tang and Zekai Zhang and Zhengyi Wang and An Yang and Bowen Yu and Chen Cheng and Dayiheng Liu and Deqing Li and Hang Zhang and Hao Meng and Hu Wei and Jingyuan Ni and Kai Chen and Kuan Cao and Liang Peng and Lin Qu and Minggang Wu and Peng Wang and Shuting Yu and Tingkun Wen and Wensen Feng and Xiaoxiao Xu and Yi Wang and Yichang Zhang and Yongqiang Zhu and Yujia Wu and Yuxuan Cai and Zenan Liu},
      year={2025},
      eprint={2508.02324},
      archivePrefix={arXiv},
      primaryClass={cs.CV},
      url={https://arxiv.org/abs/2508.02324}, 
}

\clearpage
\setcounter{page}{1}
\appendix

\section{Training Details and Hyperparameters}
\label{supp_train}

\subsection{Training Setup}
\label{supp_train_setup}

We fine-tune BAGEL-7B-MoT~\cite{deng2025bagel} using PyTorch FSDP (Fully Sharded Data Parallel) with \texttt{bf16} mixed precision on 8 NVIDIA A100 80\,GB GPUs.
Table~\ref{tab:supp_hyperparams} summarizes the key hyperparameters.

\begin{table}[h]
\centering
\caption{Training hyperparameters.}
\label{tab:supp_hyperparams}
\small
\begin{tabular}{ll}
\toprule
\textbf{Parameter} & \textbf{Value} \\
\midrule
Optimizer & AdamW ($\beta_1{=}0.9$, $\beta_2{=}0.95$, $\epsilon{=}10^{-15}$) \\
Learning rate & $1 \times 10^{-5}$ (constant after warmup) \\
Warmup steps & 2{,}000 \\
Gradient clipping & max norm $= 1.0$ \\
EMA decay & 0.9999 \\
\midrule
Max tokens per batch & 32{,}768 \\
Max tokens per sample & 24{,}576 \\
Input image resolution & $1024 \times 1024$ (PET) / $512 \times 512$ (PT, MVC) \\
IPT latent resolution & $64 \times 64$ (Latent-64) \\
\midrule
Flow-matching loss weight ($\lambda_{fm}$) & 1.0 \\
Language modeling loss weight ($\lambda_{lm}$) & 1.0 \\
\midrule
Frozen modules & VAE encoder \& decoder \\
Fine-tuned modules & LLM (all layers), ViT encoder, connector \\
\bottomrule
\end{tabular}
\end{table}

\noindent\textbf{System prompts.}
All training modes share one of two system prompts prepended to every input:

\begin{itemize}
    \item \textbf{Thinking prompt} (used for IPT, Text CoT, and label-only):

    \begin{quote}\small
    \textit{Let's think step by step to answer the question. For text-based thinking, enclose the process within \texttt{<think>} \texttt{</think>}. For visual thinking, enclose the content within \texttt{<image\_start>} \texttt{</image\_end>}. Finally conclude with the final answer wrapped in \texttt{<answer></answer>} tags.}
    \end{quote}

    \item \textbf{Answer-only prompt} (used for the answer-only portion of mixed training):

    \begin{quote}\small
    \textit{Answer the question directly. Wrap your answer in \texttt{<answer></answer>} tags. Do not think or generate any images.}
    \end{quote}
\end{itemize}

\noindent\textbf{Training modes.}
Table~\ref{tab:supp_modes} summarizes the five training configurations evaluated in this work.
Each mode differs in the output format the model is trained to produce, and correspondingly in which loss terms are active.
In IPT mode, both $\mathcal{L}_{fm}$ and $\mathcal{L}_{lm}$ are active; in Text CoT and label-only, only $\mathcal{L}_{lm}$ is used.

\begin{table}[h]
\centering
\caption{Training modes and their output formats. \texttt{[IMG]} denotes the generated intermediate image tokens.}
\label{tab:supp_modes}
\footnotesize
\setlength{\tabcolsep}{4pt}
\begin{tabular}{llp{5.8cm}}
\toprule
\textbf{Mode} & \textbf{Prompt} & \textbf{Training target} \\
\midrule
Label-only & Think & \texttt{<answer>A</answer>} \\
Text CoT & Think & \texttt{<think>}\textit{reasoning}\texttt{</think>}\newline\texttt{<answer>A</answer>} \\
IPT & Think & \texttt{<think>}\textit{task prompt}\texttt{</think>}\newline\texttt{<image\_start>[IMG]<image\_end>}\newline\texttt{<answer>A</answer>} \\
\midrule
Mixed & 50\% Think / & 50\% IPT + \\
       & 50\% Ans-only & 50\% answer-only \\
\bottomrule
\end{tabular}
\end{table}

\noindent\textbf{Mixed training.}
Mixed training combines 50\% IPT examples (with the thinking prompt and visual generation targets) and 50\% answer-only examples (with the answer-only prompt and direct answers).
The two data subsets are mixed at the dataloader level: both dataset names and sample counts are specified in the training configuration, and the dataloader interleaves batches from both sources.
The model learns to switch between generating imaginative perception tokens and producing direct answers based on which system prompt is provided, enabling a single checkpoint to operate in either mode at inference time.

\noindent\textbf{Text CoT generation.}
Text chain-of-thought training targets are generated by GPT-5.1 using ground-truth scene metadata from the simulator.
For each training example, GPT-5.1 receives the input image, the ground-truth answer, and a task-specific instruction, and produces a step-by-step textual reasoning trace (100--300 words).
Below we describe the task-specific prompts.

\paragraph{Path tracing CoT prompt.}
The system prompt instructs the model to act as a spatial reasoning AI that solves indoor navigation questions by analyzing top-down views step by step.
The task instruction is:

\begin{quote}\small
\textit{You are navigating along a numbered path through an indoor scene. The top-down view shows the path with waypoints and midpoints. You need to determine what object is visible from a specific side (left or right) at a midpoint.}

\textit{Reason step by step:
(1) Identify the path direction (which waypoint to which).
(2) Determine your orientation at the midpoint.
(3) Figure out what ``left'' or ``right'' means given that orientation.
(4) Analyze the top-down layout to identify objects on that side.
(5) Compare against the answer choices and eliminate wrong ones.}
\end{quote}

\begin{figure}[]
\centering
\includegraphics[width=\linewidth]{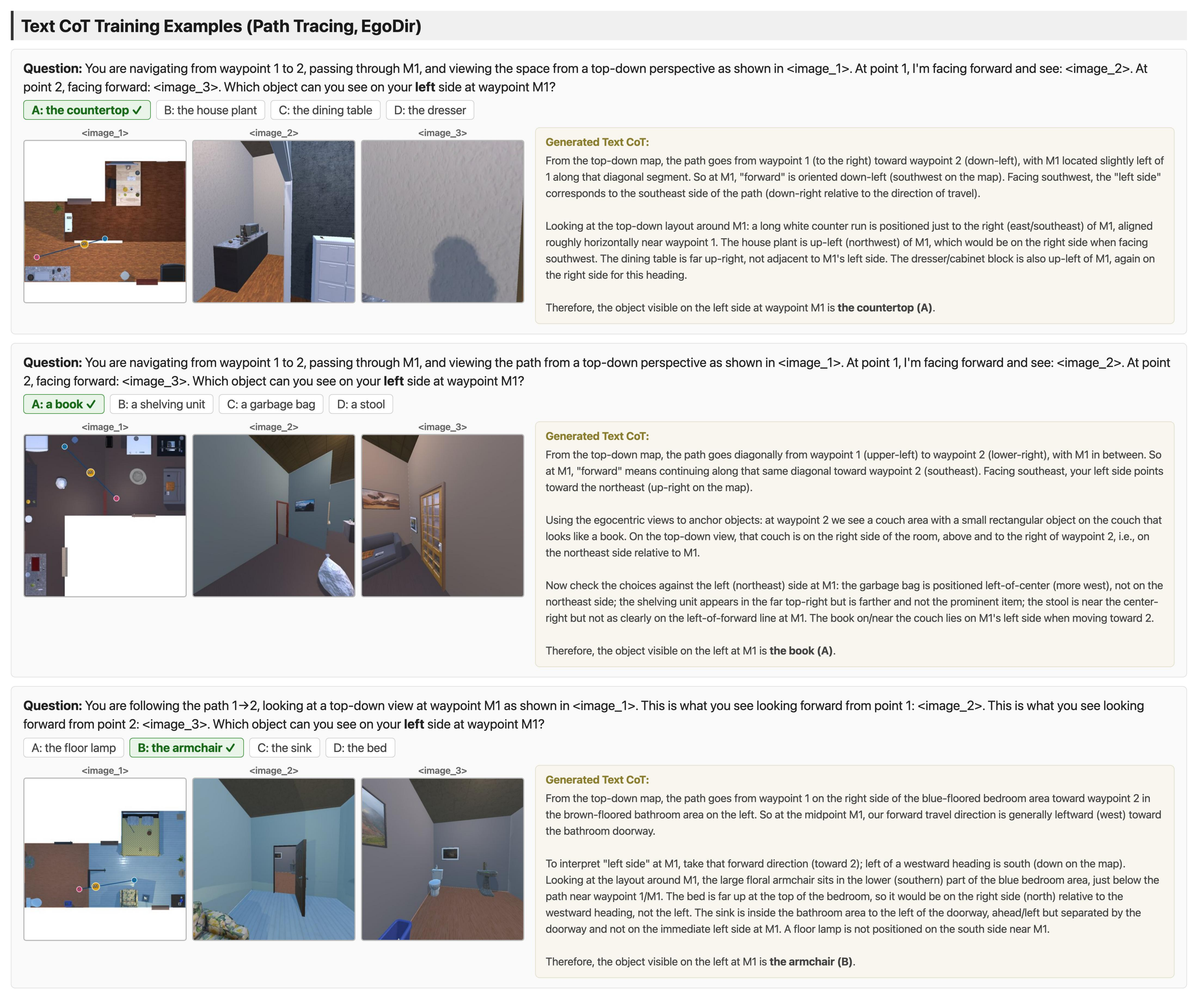}
\caption{\textbf{Text CoT training examples for path tracing (EgoDir setting).} Each example shows the input images (top-down map and egocentric views at endpoints), the question with answer choices, and the GPT-5.1-generated reasoning trace. The reasoning follows the structured prompt: identifying path direction, determining orientation at $M_1$, interpreting left/right relative to that orientation, and eliminating distractors.}
\label{fig:pt_textcot}
\end{figure}

\noindent Figure~\ref{fig:pt_textcot} shows three Text CoT training examples for path tracing in the EgoDir setting, illustrating how the generated reasoning traces analyze the top-down layout, determine the agent's orientation at the midpoint, and systematically eliminate incorrect answer choices.

\paragraph{Perspective Taking CoT prompt.}
For Perspective Taking, the model receives the input image together with privileged hidden information, the ground-truth answer and a second image showing the target view after the camera motion, to ensure correctness. The generated chain-of-thought must be written as if only the original image and question were available, without referencing any hidden information. The prompt is:
 
\begin{quote}\small\itshape
You generate student-facing chain-of-thought explanations for visual navigation/spatial reasoning from a single image. You may receive hidden privileged information (metadata, ground-truth answer, and a second image) to ensure correctness, but your explanation must be written as if you only saw the original image and question. Do not mention, quote, paraphrase, or allude to any hidden information.
 
\smallskip
\noindent Student-visible input (ONLY what the explanation may reference):
(1) Image A (original scene with a red ``X'' on the floor).
(2) Question: \{Q\}
 
\noindent Hidden privileged information (FOR CORRECTNESS ONLY --- MUST NOT APPEAR IN THE EXPLANATION):
(1) Correct answer: \{A\}
(2) Image B (ground-truth final view after moving and turning)
 
\smallskip
\noindent Task: Write a chain-of-thought that a student could produce from Image A and the question alone (not more than 10 sentences).
 
\noindent Hard constraints: Do NOT mention Image B, metadata, coordinates, angles, yaw/pitch/roll, agent states, movement info, IDs, or any numbers from the hidden block. The explanation must be plausible using only Image A: talk only about visible objects, their relative positions, and the effect of turning left/right. Keep it short (3--6 steps). No extra commentary. End with the correct final answer: \emph{My answer is [A]/[B].} Do not answer based on your own perspective; stick to the ground-truth answer provided.
 
\smallskip
\noindent Output format:\\
Step 1: Identify the object in the question and describe its relative location in the original view.\\
Step 2: Describe the walk-to-X action (translation without changing facing direction).\\
Step 3: Describe the turn action (update egocentric left/right).\\
Step 4: Conclude whether the object is on the left or right relative to the agent.
\end{quote}
 
\noindent An example of a generated thought:
\begin{quote}\small\itshape
Step 1: The oven is on the right wall of the kitchen, a bit ahead of the current viewpoint, and the red X is on the floor between the viewer and that oven wall.
Step 2: Walking straight to the X without turning keeps me facing the same direction, so I move forward toward the area in front of the oven and end up nearer that wall.
Step 3: From the X, turning right 90 degrees means I would now face the wall with the oven and cabinets.
Step 4: Since I walked closer to that wall before turning, I end up standing nearer to the oven than I was at the start.
My answer is [A].
\end{quote}
 
\paragraph{Multiview Counting CoT prompt.}
For Multiview Counting, the model receives per-frame metadata listing which object instances (by ID) are visible in each frame, and is asked to produce a frame-by-frame reasoning trace without directly referencing the hidden instance IDs. The prompt is:
 
\begin{quote}\small\itshape
You are given an object-counting question based on multiple images from the same scene. You will be provided with: the target object, the correct total count, and frame-level information listing which object instances are visible in each frame (this information is hidden and should not be referenced directly).
 
\smallskip
\noindent IMPORTANT: An empty list for a frame means that no objects of the target type are visible in that frame.
 
\smallskip
\noindent Write a brief, frame-by-frame explanation describing what is visible in each frame. Do not mention object IDs or refer to them explicitly. When explaining each frame, do not count objects that were already visible in previous frames.  After the frame-by-frame explanation, conclude with: ``The total number is X.''
 
\smallskip
\noindent Object: \{O\}\\
Correct total count: \{answer\}\\
Frames: \{frames\_text\}
\end{quote}

 \noindent An example of a generated thought:
\begin{quote}\small\itshape
Frame 1: No bowls are visible.  Frame 2: A bowl appears and is visible for the first time.  Frame 3: No new bowls appear; the same bowl from before may still be present. The total number is 1.
\end{quote}

\subsection{Evaluation Setup}
\label{supp_eval}

\noindent\textbf{Inference parameters.}
Table~\ref{tab:supp_eval_params} lists the inference hyperparameters used across all evaluations.

\begin{table}[h]
\centering
\caption{Inference hyperparameters for evaluation.}
\label{tab:supp_eval_params}
\small
\begin{tabular}{ll}
\toprule
\textbf{Parameter} & \textbf{Value} \\
\midrule
Text temperature & 0.3 \\
Sampling & do\_sample = True \\
Max thinking tokens & 4{,}096 \\
\midrule
Diffusion timesteps & 50 \\
Timestep shift & 3.0 \\
Text CFG scale & 4.0 \\
Image CFG scale & 2.0 \\
CFG interval & $[0.0, 1.0]$ \\
Generated image resolution & $1024 \times 1024$ \\
Max generation rounds & 1 \\
\midrule
GPU allocation & 2$\times$ A100 80\,GB (model parallelism) \\
\bottomrule
\end{tabular}
\end{table}

\noindent\textbf{Evaluation modes.}
At inference, each model variant is evaluated in the mode that matches its training configuration.
IPT models are evaluated in two settings: (1)~\emph{imagination mode}, where the model generates an intermediate image before answering, and (2)~\emph{answer-only mode}, where the model produces only a text answer without generating any image.
For models trained with visual generation (IPT, Mixed), the VAE weights are always loaded (\texttt{visual\_gen=True}), and input images are encoded through both the ViT and VAE pathways (\texttt{vae\_input=True}) to match the training-time encoding.
Without setting \texttt{vae\_input=True}, a train--eval mismatch would occur: during training, input images pass through both VAE and ViT, but the default evaluation behavior sends inputs through ViT only.\footnote{For Path-Tracing, we found that setting \texttt{vae\_input=False} actually improves generalization to real environments.}

\noindent\textbf{Answer extraction.}
We extract the predicted answer letter from model outputs using a cascading rule-based procedure:
(1)~parse \texttt{<answer>X</answer>} tags;
(2)~extract from \texttt{\textbackslash boxed\{X\}} format;
(3)~match patterns such as ``the answer is X'';
(4)~detect bold letter formatting (\texttt{**X**});
(5)~fall back to the last single letter in the response.
All benchmarks use the same unified scoring function that compares the extracted letter against the ground-truth answer.

\section{Data Curation Details}
\label{supp_data}

\subsection{Path Tracing}
\label{supp_data_pt}

We generate path tracing data from two sources: AI2-THOR (synthetic) and Matterport3D (real-world).

\subsubsection{AI2-THOR}

\paragraph{Scene selection.}
We use 120 standard iTHOR~\cite{kolve2017ai2} scenes spanning four room types (kitchens, living rooms, bedrooms, bathrooms), with 30 scenes per type, split into train (20), val (5), and test (5) per type.
The training set additionally incorporates procedurally generated houses from ProcTHOR-10k~\cite{deitke2022procthor}, which include a fifth room type (hallways, offices, dining rooms).

\paragraph{Path sampling.}
For each scene, we sample feasible two-waypoint paths on the navigation mesh, balanced across room types and three distance bins: short ($1$--$2$\,m), medium ($2$--$4$\,m), and long (${\geq}4$\,m).
Grid-based path sampling uses a spacing of $0.5$\,m with a minimum waypoint separation of $1.0$\,m.

\paragraph{Camera configuration.}
All views are rendered at $1024 \times 1024$ resolution.
To increase viewpoint diversity, we randomize the camera height (sampled from seven values between $1.4$ and $1.8$\,m), field of view ($75°$--$120°$), and pitch ($-5°$ to $5°$).

\paragraph{Rendering.}
At each sample we render top-down views, egocentric forward views at both endpoints, and a sweep of candidate sideviews at the midpoint $M_1$.
The sideview sweep covers $7$ yaw angles $\times$ $7$ horizontal offsets $\times$ $3$ pitch values, yielding $147$ candidate views per midpoint.
We select the sideview that best exposes the queried object using simulator segmentation masks, requiring a minimum object coverage of $0.15\%$ of the image area and a maximum view angle of $90°$ relative to the path direction.

\paragraph{Question generation.}
Questions are generated from templates (``Which object can you see on your \{side\} at waypoint M1?'') with four choices: the correct answer drawn from verified visible objects and three distractors drawn from the opposite side or a global object pool.
Each base MCQ is expanded into eight input variants by combining different image types (top-down path, top-down with arrow, top-down with midpoint marker, dollhouse view) and egocentric cue availability (with/without endpoint views).

\paragraph{TIFA filtering.}
We apply TIFA-style filtering~\cite{hu2023tifa} to ensure question quality.
Each candidate is decomposed into binary visibility queries and verified by GPT-4.1 with three-round majority voting, using early exit after round 2 when unanimous.
Samples are dropped if (1) the correct answer is not visible in the sideview, (2) a distractor is also visible in the sideview, or (3) the model answers incorrectly even when provided the sideview.
We further remove samples where GPT-4.1 answers correctly from the top-down view and egocentric endpoint views alone, ensuring the benchmark requires genuine spatial imagination.

\paragraph{Debiasing.}
Answer choices are reshuffled with per-sample deterministic seeds to remove positional bias.
We additionally verify that per-object and per-room answer distributions remain approximately uniform.

\paragraph{Statistics.}
The synthetic training set contains $11{,}204$ examples.

\subsubsection{Real-World Data (Matterport3D)}

To evaluate cross-domain transfer, we construct a real-world test set from Matterport3D~\cite{chang2017matterport3d} top-down views.

\paragraph{Image collection.}
We collect top-down screenshots from Matterport 3D indoor tours, capturing per-floor views with UI elements removed and dark borders cropped.

\paragraph{Auto-annotation.}
We annotate walking paths on each image using a two-pass GPT pipeline.
In the first pass, GPT proposes $N$ candidate walking paths, each defined by three waypoints (start ``1'', midpoint ``M1'', end ``2'') placed on open floor areas.
In the second pass, the proposed waypoints are drawn on the image and GPT is asked to (a) verify that $M_1$ lies on walkable floor, adjusting its position if necessary, and (b) identify $2$--$5$ visible furniture items or objects on each side (left/right) of the path at $M_1$.

\paragraph{Post-processing.}
Several geometric filters are applied to ensure path quality.
Waypoints are clamped to image bounds, and paths shorter than $30\%$ of the shorter image dimension are rejected.
$M_1$ is snapped onto the line segment between waypoints 1 and 2 and constrained to the $[0.2, 0.8]$ interval to avoid proximity to endpoints.
Paths that traverse dark or background regions (more than $20\%$ dark pixels along the path) are discarded.

\paragraph{TIFA filtering.}
Because no side-view images exist for real environments, TIFA verification operates on the top-down image only.
We verify three properties: (1) all three waypoints lie on walkable floor (not on furniture, walls, or background); (2) each annotated object is visible in the image; and (3) each object is on the correct side of the path.
Paths with fewer than $2$ verified objects on either side are dropped.
Majority voting across up to $3$ rounds is used, with early exit when the first two rounds agree.

\paragraph{Human review.}
After automated filtering, all surviving annotations undergo human review, where annotators can approve, delete, or edit individual paths and their associated object lists.

\paragraph{Question generation.}
Each verified path yields eight question variants (forward/reverse $\times$ left/right $\times$ arrow/no-arrow).
When the walking direction is reversed ($2 \to 1$), left and right swap because the agent faces the opposite direction.
Distractors are drawn from opposite-side objects first, then supplemented from a global pool of $50$ common indoor objects.
Answer choices are shuffled with per-sample deterministic seeds to eliminate positional bias.
Because the real environments lack egocentric viewpoints, we evaluate on the Path and PathArr settings only.
The real-world benchmark contains $332$ human-verified questions.

\subsection{Perspective Taking}
\label{supp_data_pet}

We generate perspective taking data from three sources: AI2-THOR (synthetic), Habitat (photorealistic scans), and Visual Spatial Tuning (real-world images).
All sources share the same task structure---given a first-person view with a marked target position, answer a spatial question about the scene from the new viewpoint---but differ in visual domain and 3D engine.

\subsubsection{AI2-THOR}

\paragraph{Scene selection.}
We use procedurally generated indoor scenes from ProcTHOR~\cite{deitke2022procthor}, which provides diverse house layouts with varied room configurations.
For each scene, we sample multiple camera positions from the navigable area and generate perspective-taking examples at each position.

\noindent\textbf{Target position placement.}
The target position (marked with a red ``X'' on the input image) is determined by raycasting from the camera into the scene.
To ensure physically plausible movement targets, we filter ray hits to \emph{ground-only} surfaces: floors, carpets, rugs, and tiles.
Hits on furniture, tabletops, or other elevated surfaces are rejected.
The hit point must lie within $\pm0.2$\,m of the ground-level height.
Up to 40 raycasting attempts are made per sample; if no valid ground hit is found, the sample is skipped.

\noindent\textbf{Viewpoint transformation.}
The agent is teleported to the target position and rotated $90°$ in a randomly chosen direction (left or right with equal probability), simulating a realistic movement-and-turn action.
A ground-truth novel-viewpoint image is rendered at this new pose to serve as the imaginative perception target.

\noindent\textbf{Object filtering.}
To ensure well-defined questions, target objects must satisfy several criteria:
\begin{itemize}
    \item Visible in \emph{both} the original and new viewpoint (dual-view visibility).
    \item Occupy at least 0.4\% of the image area.
    \item Lie within 5\,m of the camera.
    \item Fall at least 150\,px from the image edge (to avoid partially visible objects).
    \item For relative position questions: the object must be unambiguously on one side of the image center, with a 150\,px margin from the center line.
\end{itemize}
Additionally, we enforce a \emph{left-right eligibility} constraint: an object is only used for relative position questions if it appears on at most one side of the image (not straddling the center), ensuring that the left/right answer is unambiguous.

\noindent\textbf{Question generation.}
We generate two types of questions with 10 template variants each, varying in person perspective (first, second, third person) and formality level:
\begin{itemize}
    \item \textbf{Distance change}: ``After moving to `X' and turning \{direction\} $90°$, will the \{object\} get closer or further?'' Requires a minimum distance change of $\pm0.5$\,m between the old and new camera positions.
    \item \textbf{Relative position}: ``After moving to `X' and turning \{direction\} $90°$, will the \{object\} be on your left or right?'' Left/right is determined by the object's 2D position in the new-viewpoint image.
\end{itemize}

\noindent\textbf{Sub-categories.}
As described in \cref{sec:perspective_taking}, the six balanced sub-categories arise from the combination of question type and answer:
two for distance change (\emph{closer}, \emph{further}) and four for relative position (\emph{left$\to$left}, \emph{left$\to$right}, \emph{right$\to$left}, \emph{right$\to$right}), where the notation indicates the object's lateral position before and after the viewpoint transformation.
The training set is balanced across all six sub-categories.

\noindent\textbf{Image annotation.}
Each input image is annotated in two versions: (1)~with only a red ``X'' marking the target position, and (2)~with the ``X'' plus a blue directional arrow indicating the agent's facing direction after rotation.
The arrow version provides an additional spatial cue at evaluation time.

\paragraph{Statistics.}
The AI2-THOR training set contains $20{,}531$ examples across 98 scenes, with an average of ${\sim}210$ questions per scene.

\subsubsection{Habitat}

\noindent\textbf{Scene source.}
We use photorealistic 3D scans from HM3D (Habitat-Matterport 3D)~\cite{chang2017matterport3d} with semantic annotations.
Only single-floor scenes are selected (floor-level Y-variance $< 2.0$\,m) to avoid cross-level ambiguities.

\paragraph{Camera configuration.}
Images are rendered at $1024 \times 1024$ resolution with a horizontal field of view of $90°$.
The sensor height is set to $1.25$\,m above the navigable floor surface, matching a standing human eye level.

\noindent\textbf{Target position and viewpoint.}
Camera~A (original viewpoint) is placed at a random navigable point with a random yaw.
The target position (Camera~B) is determined by selecting a visible object as an anchor: Camera~B is placed at the object's XZ coordinates, offset slightly along the facing direction to avoid clipping into geometry, and snapped to the nearest navigable point on the mesh (within a $1.0$\,m snap radius).
The ground-truth novel-viewpoint image is rendered at Camera~B's position and orientation.

\noindent\textbf{Object filtering.}
We apply a strict whitelist of ${\sim}70$ mainstream furniture categories (seating, tables, beds, storage, appliances, bathroom fixtures) and exclude structural elements (walls, floors, doors).
Objects must occupy at least 0.8\% of the image area in the original frame and at least 0.5\% in the imagined frame.
To avoid ambiguous references, only objects whose category is \emph{unique} in the frame are used (\eg, if two chairs are visible, neither is selected as a question target).
An edge margin of 200\,px is applied.

\noindent\textbf{Left/right determination.}
Object laterality is determined by the mean x-coordinate of the object's semantic segmentation mask in the rendered image, with a 180\,px margin from the image center ($512$\,px).
Objects falling in the center zone ($332 < x < 692$) are excluded as ambiguous.

\paragraph{Statistics.}
The Habitat training set contains $19{,}998$ examples balanced across the six sub-categories.

\subsubsection{Real-World Data (VST)}


To bridge the synthetic-to-real domain gap, the \emph{mixed} training variant incorporates $15{,}000$ real-world examples drawn from the camera motion subset of the Visual Spatial Tuning (VST) dataset~\cite{yang2025visual}. Each example contains a pair of multi-view images captured from different viewpoints in real indoor scenes, along with a question about the camera motion between them and a corresponding answer.
 
\paragraph{Filtering uncertain answers.}
We first filter out examples whose answers are uncertain or underspecified using GPT-5.1, prompting it as a binary classifier. Any example whose answer contains phrases such as ``cannot be determined,'' ``unknown,'' or ``insufficient information'' is removed. The filtering prompt is:
 
\begin{quote}
\itshape
You are a binary classifier.\\
Proposed Answer:\\
\textless start\_answer\textgreater\{A\}\textless end\_answer\textgreater\\
If the answer contains ANY of the following phrases or meanings, output ``NO'':
\begin{itemize}
    \item cannot be determined
    \item insufficient information
    \item not enough information
    \item unknown
    \item unclear
    \item cannot tell
    \item impossible to determine
    \item indeterminate
\end{itemize}
Only output ``YES'' if the answer clearly states a specific, determined choice (e.g., a direction, location, label, or concrete option).\\
Output exactly one token: YES or NO.
\end{quote}
 
\paragraph{Rewriting into generation prompts.}
After filtering, we use GPT-5.1 to rewrite each question, answer pair into a generation prompt describing the camera motion \emph{from the first image to the second}. This requires careful handling of reference frame direction: if the original question asks where the first camera is relative to the second image, the motion direction must be inverted before constructing the prompt. The rewriting prompt is:
 
\begin{quote}
\itshape
You are given a question about camera motion between two images and its correct answer.\\[4pt]
IMPORTANT INVERSION RULE:
\begin{itemize}
    \item If the question asks ``where is the FIRST camera relative to the SECOND image'' (or uses the second image as reference), then the answer describes motion FROM second TO first.
    \item In this case, you MUST invert the direction to get motion FROM first TO second.
    \item If the question asks ``where is the SECOND camera relative to the FIRST image'' (or uses the first image as reference), NO inversion is needed.
\end{itemize}
STEPS:
\begin{enumerate}
    \item Determine which image is the reference point in the question.
    \item If the reference is the second image, invert the direction in the answer.
    \item Using the final motion FROM first TO second, create a generation prompt.
\end{enumerate}
INVERSION EXAMPLES:
\begin{itemize}
    \item ``right'' $\to$ ``left''
    \item ``left'' $\to$ ``right''
    \item ``front'' $\to$ ``back''
    \item ``back'' $\to$ ``front''
    \item ``front left'' $\to$ ``back right''
    \item ``back right'' $\to$ ``front left''
\end{itemize}
Your output should start with ``generate'' and describe viewing the scene from the new camera position after applying the motion from the first image.\\[4pt]
Question: \textless start\_question\textgreater\{Q\}\textless end\_question\textgreater\\
Answer: \textless start\_answer\textgreater\{A\}\textless end\_answer\textgreater\\[4pt]
First, identify the reference image. Then apply inversion if needed. Then generate your output.
\end{quote}
 
The resulting generation prompts condition the model on the first view and the inferred camera motion, with the second view serving as the imaginative perception target. Because no programmatic 3D annotation is available for these real scenes, this data serves as a domain bridge rather than a source of the full six-sub-category question format.

\noindent\textbf{Mixed training composition.}
The mixed PET training variant combines AI2-THOR ($20{,}531$), Habitat ($19{,}998$), and VST ($15{,}000$) examples, totaling $55{,}529$ samples.

\subsection{Multiview Counting}
\label{supp_data_mvc}

We construct multiview counting data from both synthetic and real-image sources.
Our main training set is generated from ProcTHOR/AI2-THOR environments, which provide full 3D supervision for both egocentric observations and top-down bird's-eye-view (BEV) targets.
To complement this synthetic source, we additionally curate two real-image multiview counting sets from MessyTable and ScanNet++, which expose the model to real visual appearance and partial observability under natural image statistics.

\subsubsection{ProcTHOR / AI2-THOR}

We generate the main multiview counting training set from AI2-THOR environments, using two trajectory types that capture complementary modes of partial observability.

\paragraph{Trajectory types.}
\begin{itemize}
  \item \textbf{Rotation}: The agent remains at a fixed position and rotates in $90^\circ$ increments through four cardinal directions ($0^\circ$, $90^\circ$, $180^\circ$, $270^\circ$), producing four frames that together cover a $360^\circ$ panorama of the surrounding area.
  \item \textbf{Multi-camera}: The agent traverses a square path, capturing one frame at each of four corners. This setup simulates a multi-camera rig where viewpoints are spatially distributed around the scene.
\end{itemize}
Both trajectory types produce exactly four input frames per sample.

\paragraph{Bird's-eye view (BEV) generation.}
The ground-truth intermediate image is a top-down BEV map rendered from an overhead camera in the 3D scene.
To ensure the BEV only covers the \emph{explored} area (the region visible from the input frames), we crop the map with trajectory-aware padding: $5$\,m around the agent position for rotation trajectories and $4$\,m around the traversed path for multi-camera trajectories.
Object counts in the cropped BEV are validated against the segmentation maps to ensure consistency.

\paragraph{Object filtering and category balancing.}
Structural elements (walls, floors, ceilings, doorways) are excluded from counting targets.
Target objects must be visible in both the first-person frames and the cropped top-down segmentation map (with a minimum coverage of $0.1\%$).
Because initial generation heavily favors count$=1$ questions (${\sim}82\%$), we apply iterative rebalancing: high-frequency categories are capped at $9.9\%$ of the dataset, and count$=1$ samples are downsampled.
This produces a more uniform distribution across object categories and count values.

\paragraph{Question format and distractor generation.}
Questions follow the template: ``How many \{category\}(s) are in this area?'' with four answer choices (A--D).
Distractors are sampled from $\pm1$ and $\pm2$ of the correct count, producing plausible alternatives.
Negative counts are removed, and the four options are shuffled with a per-sample deterministic seed to eliminate positional bias.

\paragraph{Statistics.}
The synthetic training set contains $17{,}079$ examples generated from ProcTHOR~\cite{deitke2022procthor} scenes, covering both trajectory types.

\subsubsection{MessyTable}

To expose the model to real tabletop imagery with severe clutter and occlusion, we construct an additional multiview counting set from MessyTable~\cite{cai2020messytable}. Each scene contains multiple camera views of the same tabletop arrangement together with instance-level annotations.

\paragraph{Scene-level counting targets.}
For each scene, we aggregate annotations across all cameras and de-duplicate instance IDs across views, so that the ground-truth answer corresponds to the number of unique physical objects rather than the sum of per-view detections. Counting targets are defined at the subclass level and mapped
to readable category names.

\paragraph{Target and view sampling.}
For each scene, we sample one target category from the categories present in the scene. To reduce the dominance of trivial singleton cases, sampling is biased toward categories with count $\geq 2$: when both singleton and multi-instance categories are available, $90\%$ of samples are drawn from
the multi-instance bucket and $10\%$ from the singleton bucket. Input images are selected from the eight surrounding non-top cameras, with priority given to views in which the target category is absent. When too few such views exist, they are supplemented with additional non-adjacent views to
maintain viewpoint diversity. This makes the final count require aggregation across multiple views rather than inspection of a single image.

\paragraph{Top-view supervision and question generation.}
Each sample also stores the canonical top-view image as the reasoning target. In the exported JSONL format, image inputs use centered crops derived from the union of all annotated object boxes in each selected camera view, which reduces empty borders while preserving the visible object layout.
Questions are instantiated from a diverse pool of natural-language counting templates such as ``How many \{object\} are in this scene?''

\subsubsection{ScanNet++}

We further construct a real indoor multiview counting set from ScanNet++~\cite{yeshwanth2023scannet++}, using iPhone image trajectories paired with labeled 3D scene reconstructions. Compared with MessyTable, this set covers larger indoor spaces, more varied viewpoints, and more realistic household layouts.

\paragraph{Top-down map and candidate view generation.}
For each scene, we first generate a top-down map from the labeled 3D reconstruction. Because raw point-cloud renderings are visually sparse and unrealistic, we further use Qwen-Image-Edit~\cite{wu2025qwenimagetechnicalreport} to transform the rendered top-down visualization into a more realistic top-down image while preserving the scene layout. We then build a candidate egocentric view pool by combining a small set of canonical iPhone views with additional randomly sampled frames, requiring each extra frame to differ from the canonical views by at least a minimum yaw angle.

\paragraph{Visibility estimation and target selection.}
We estimate which object instances are visible in each candidate frame by projecting the labeled 3D scene into the camera views using mesh ray-casting, and use the semantic annotations to obtain scene-level category counts. Top-down maps are filtered by automatic quality rules to remove blurry,
blank, or low-texture renderings. Candidate counting targets are restricted to non-structural object categories with bounded scene-level counts and sufficient visible support in the candidate views. To avoid metadata leakage, target selection is performed in a blind setting: the model is shown
the top-down image and candidate labels, but not their annotated counts, and we keep only categories whose visually predicted count matches the ground truth. When multiple valid categories remain, we preferentially sample categories with counts greater than one.

\paragraph{Final evidence image selection.}
The final evidence set is selected in two stages. We first greedily choose images that jointly cover all instances of the target category while enforcing a minimum yaw-separation constraint. We then fill the remaining slots with views that add new foreground content and viewpoint diversity.
Each final sample contains $5$--$8$ egocentric images together with the top-down map as the reasoning image, and the question is rewritten into a natural counting form.

\section{Comparison with Related Spatial Benchmarks}
\label{app:benchmark_comparison}

IPT uniquely combines multiple components that existing spatial reasoning benchmarks lack. 
While prior work has focused on specific aspects—such as evaluation metrics, real-world 
data collection, or thought reasoning—IPT simultaneously provides training data, a scalable 
generation pipeline, paired ground-truth visual and textual thoughts, and a complete model 
training framework. Table~\ref{tab:benchmark_comparison} summarizes how IPT compares to 
related work across these key dimensions.

\begin{table}
\centering
\caption{Comparison of IPT with related spatial reasoning benchmarks.}
\label{tab:benchmark_comparison}
\scriptsize
\setlength{\tabcolsep}{4pt}
\begin{tabular}{lcccccc}
\toprule
& \textbf{Ours} & \textbf{SpatialViz}~\cite{wang2025spatialviz} & \textbf{SITE}~\cite{wang2025site} & \textbf{OmniSpatial}~\cite{jia2025omnispatial} & \textbf{SpatialScore}~\cite{wu2025spatialscore} & \textbf{MMSI}~\cite{yang2025mmsibench} \\
\midrule
Training data        & \checkmark & $\times$ & $\times$ & $\times$ & \checkmark & $\times$ \\
Scalable gen.        & \checkmark & $\times$ & $\times$ & $\times$ & \checkmark & $\times$ \\
GT visual thoughts   & \checkmark & $\times$ & $\times$ & partial  & $\times$   & $\times$ \\
GT text thoughts     & \checkmark & $\times$ & $\times$ & $\times$ & $\times$   & \checkmark \\
Model framework      & \checkmark & $\times$ & $\times$ & $\times$ & \checkmark & $\times$ \\
Real-world samples   & \checkmark & $\times$ & \checkmark & \checkmark & \checkmark & \checkmark \\
\bottomrule
\end{tabular}
\end{table}





\section{Additional Results}
\label{supp_additional_results}

Table~\ref{tab:main_results} in the main paper reports path tracing accuracy averaged across input settings.
Table~\ref{tab:pt_breakdown} provides the full per-split breakdown for both AI2-THOR (EgoDir, Path, PathArr) and different-environment (Real, Real+Arr) benchmarks.

\begin{table}[]
  \caption{\textbf{Path tracing per-split results.} Accuracy (\%) broken down by input setting. The main paper reports the average across these splits. For our models, accuracy reports the maximum between answer-only and free-generation inference. Best per group in \textbf{bold}.}
  \centering
  \small
  \begin{tabular}{l ccc cc}
    \toprule
    & \multicolumn{3}{c}{\textbf{AI2-THOR}} & \multicolumn{2}{c}{\textbf{Different Env.}} \\
    \cmidrule(lr){2-4} \cmidrule(lr){5-6}
    \textbf{Model}
      & EgoDir & Path & PathArr
      & Real & Real+Arr \\
    \midrule
    \multicolumn{6}{l}{\emph{VQA Models}} \\
    GPT-5            & \textbf{61.1} & \textbf{56.8} & \textbf{62.6} & \textbf{74.5} & 87.3 \\
    GPT-5.2          & 22.1 & 40.2 & 36.3 & 57.5 & 68.4 \\
    Gemini 2.5 Flash & 45.1 & 37.9 & 41.5 & 58.0 & 84.8 \\
    Gemini 3 Flash   & 48.7 & 39.1 & 39.2 & 70.1 & \textbf{96.2} \\
    InternVL3.5-8B   & 43.4 & 29.0 & 35.1 & 45.4 & 49.4 \\
    Qwen2.5-VL-7B   & 44.2 & 32.5 & 35.1 & 47.1 & 42.4 \\
    Qwen3-VL-8B     & 31.9 & 33.7 & 42.1 & 52.9 & 75.3 \\
    \midrule
    \multicolumn{6}{l}{\emph{Unified Models}} \\
    Janus-Pro-7B     & 36.3 & 30.2 & 33.9 & 34.5 & 36.1 \\
    Chameleon 7B     & 5.3 & 23.7 & 19.9 & 23.0 & 25.9 \\
    \midrule
    \multicolumn{6}{l}{\emph{Ours (fine-tuned BAGEL)}} \\
    Bagel (base)              & 36.3 & 26.0 & 27.5 & 39.7 & 45.6 \\
    Bagel (label-only)        & \textbf{73.5} & 61.5 & 62.0 & 46.6 & 62.7 \\
    + Text CoT                 & 53.1 & 47.9 & 48.0 & \textbf{52.3} & 51.3 \\
    + IPT                       & 61.1 & 43.2 & 42.7 & 46.6 & \textbf{68.4} \\
    + Mixed Training            & 71.7 & \textbf{65.1} & \textbf{63.2} & 50.6 & 66.5 \\
    \bottomrule
  \end{tabular}
  \label{tab:pt_breakdown}
\end{table}

\section{Additional Baseline Models}
\label{app:additional_baselines}

\subsection{Model Selection}
We report results for \textbf{Show-o2 7B}~\cite{xie2025showo2} and \textbf{ThinkMorph 7B}~\cite{gu2025thinkmorph}, a variant of Bagel 
with alternative fine-tuning configurations, alongside several additional baselines to 
provide comprehensive evaluation coverage.

\subsection{Comprehensive Baseline Comparison}
In addition to the primary baselines, we evaluate several variants:
\begin{itemize}
    \item \textit{Human baseline}: Expert performance on each task
    \item \textit{Random baseline}: Chance-level performance
    \item \textit{Bagel (text-only)}: Bagel without access to image prompts
\end{itemize}
Results across all baselines are presented in Table~\ref{tab:additional_baselines}.

\begin{table}
\centering
\caption{Performance comparison across baseline models and variants.}
\label{tab:additional_baselines}
\footnotesize
\setlength{\tabcolsep}{4pt}
\begin{tabular}{lccc}
\toprule
\textbf{Model} & \textbf{PT.EgoDir} & \textbf{MVC} & \textbf{Hab.\ PET} \\
\midrule
Human                        & 80.3\% & 73.8\%     & 96.3\%   \\
\midrule
Random                       & 25.0\% & 25.0\% & 50.0\% \\
Bagel (text-only)            & 15.0\%    & 28.8\%   & 51.1\%   \\
ThinkMorph                   & 40.7\% & 38.8\% & 46.6\% \\
Show-o2 7B                   & 51.0\% & 35.4\% & 48.6\% \\
\textbf{IPT (ours)}          & \textbf{61.1\%} & \textbf{67.3\%} & \textbf{87.0\%} \\
\bottomrule
\end{tabular}
\end{table}

\section{Visualizations}
\label{supp_vis}

\subsection{Path Tracing}

\begin{figure}[]
\centering
\includegraphics[width=.95\linewidth]{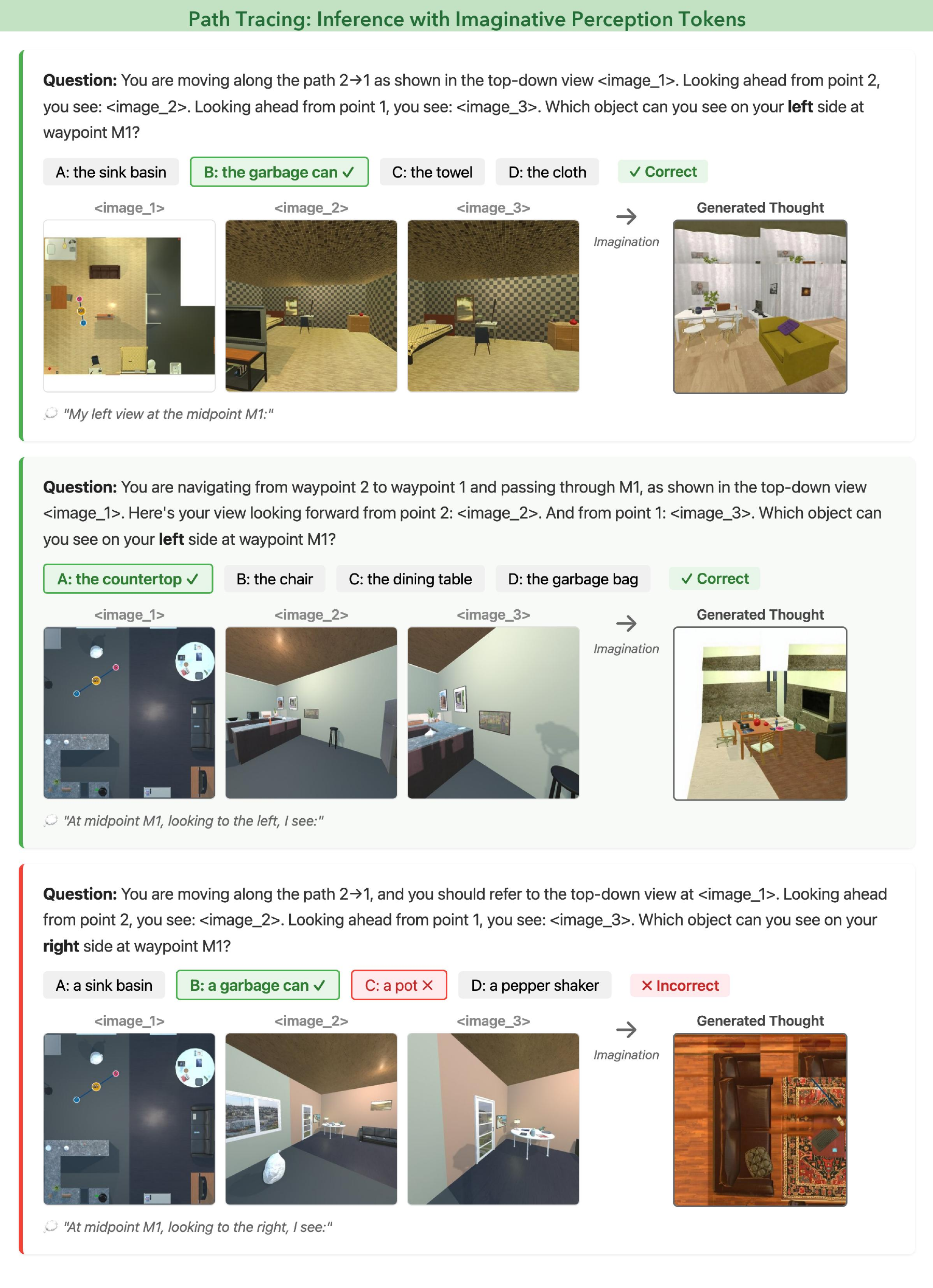}
\caption{\textbf{Path tracing with imaginative perception tokens (EgoDir setting).} The model receives a top-down map (\texttt{<image\_1>}) and egocentric views at the two endpoints (\texttt{<image\_2>}, \texttt{<image\_3>}), generates a visual thought (imagined sideview at $M_1$), and predicts an answer. Although the generated thoughts exhibit spatial imprecision and artifacts, the model still arrives at the correct answer in the first two examples, suggesting that imagination training encourages internalized spatial reasoning rather than reliance on pixel-accurate intermediate outputs. The third row shows a failure case. Correct answers are highlighted in green; incorrect predictions in red.}
\label{fig:pt_viz}
\end{figure}

\paragraph{Inference with imaginative perception.}
Figure~\ref{fig:pt_viz} shows examples of path tracing inference with imaginative perception tokens in the EgoDir setting.
Path tracing presents a particularly challenging imagination target: the model must synthesize a first-person sideview at midpoint $M_1$ from a top-down map and two egocentric endpoint views, requiring accurate reasoning about camera height, 3D object layout, and occlusion from a bird's-eye representation.
As shown in the figure, the generated visual thoughts are often spatially imprecise, with noticeable artifacts and layout errors compared to ground-truth sideviews.
Despite this, the model frequently arrives at the correct answer (rows 1 and 2).
This observation suggests that the value of imaginative perception training lies not in producing pixel-accurate intermediate images, but in encouraging the model to internalize spatial reasoning during training.
The imagination supervision acts as an auxiliary signal that shapes the model's internal spatial representations, enabling it to reason about 3D visibility even when the externalized image is imperfect.
This is further supported by the finding in our main experiments that IPT-trained models achieve strong performance in answer-only mode, where no image is generated at inference time.

\begin{figure}[]
\centering
\includegraphics[width=\linewidth]{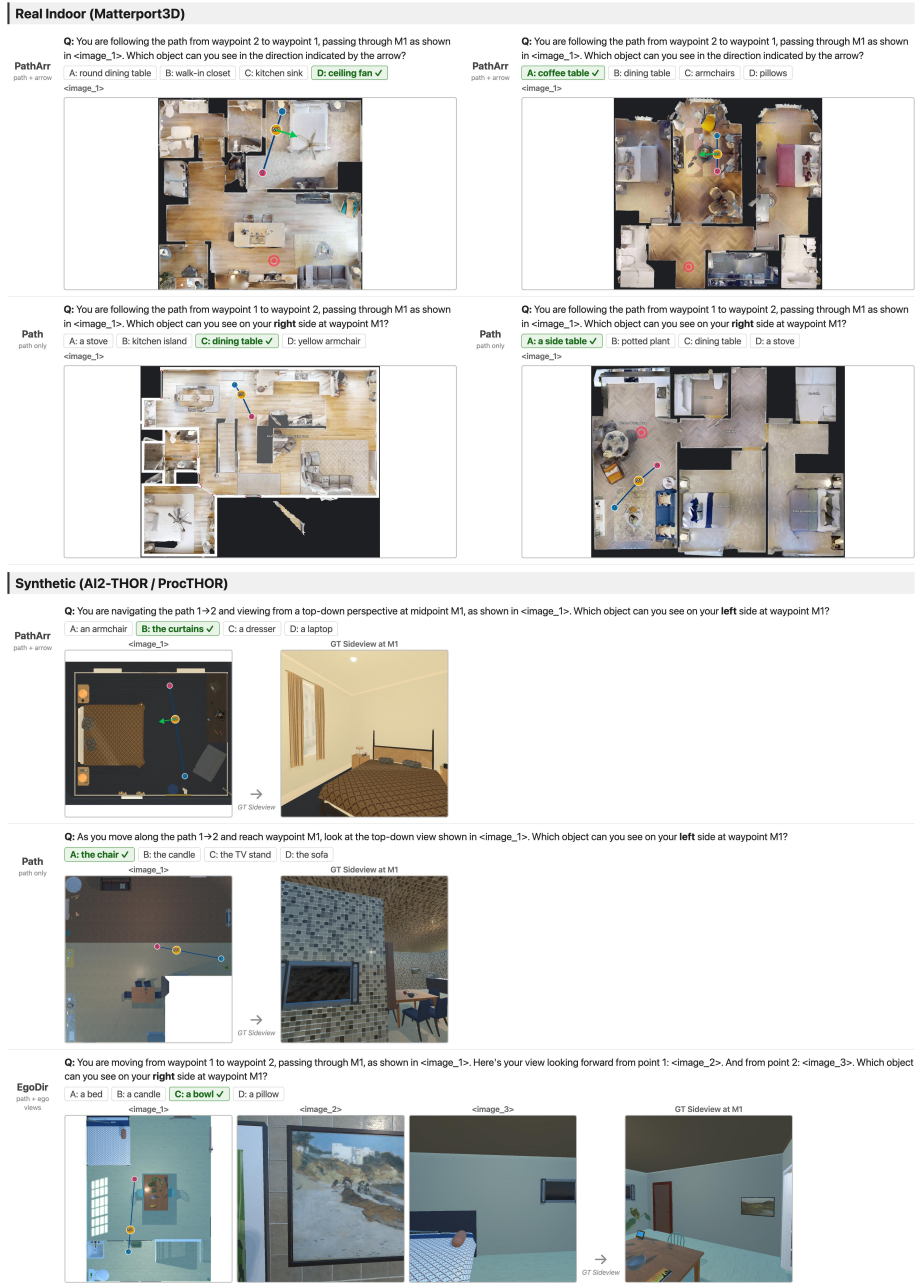}
\caption{\textbf{Path tracing dataset examples.} Top: real-world examples from Matterport3D in the PathArr and Path settings. Bottom: synthetic examples from AI2-THOR/ProcTHOR in the PathArr, Path, and EgoDir settings, with ground-truth sideviews at midpoint $M_1$ shown on the right. Each example shows the input image(s), question, and four answer choices with the correct answer highlighted.}
\label{fig:pt_dataset_examples}
\end{figure}

\paragraph{Dataset examples.}
Figure~\ref{fig:pt_dataset_examples} shows representative examples from both the real-world (Matterport3D) and synthetic (AI2-THOR/ProcTHOR) path tracing datasets across all input settings.
The real-world examples (top) use photographic top-down views and are evaluated in the Path and PathArr settings only, since real environments do not provide egocentric viewpoints at arbitrary positions.
The synthetic examples (bottom) include ground-truth sideview images rendered at the midpoint $M_1$, which serve as the imaginative perception target during training.
The PathArr setting provides a directional arrow at $M_1$ indicating the query direction, while the EgoDir setting provides egocentric forward views at both endpoints in addition to the top-down map.

\subsection{Perspective Taking}

\begin{figure}[]
\centering
\includegraphics[width=.95\linewidth]{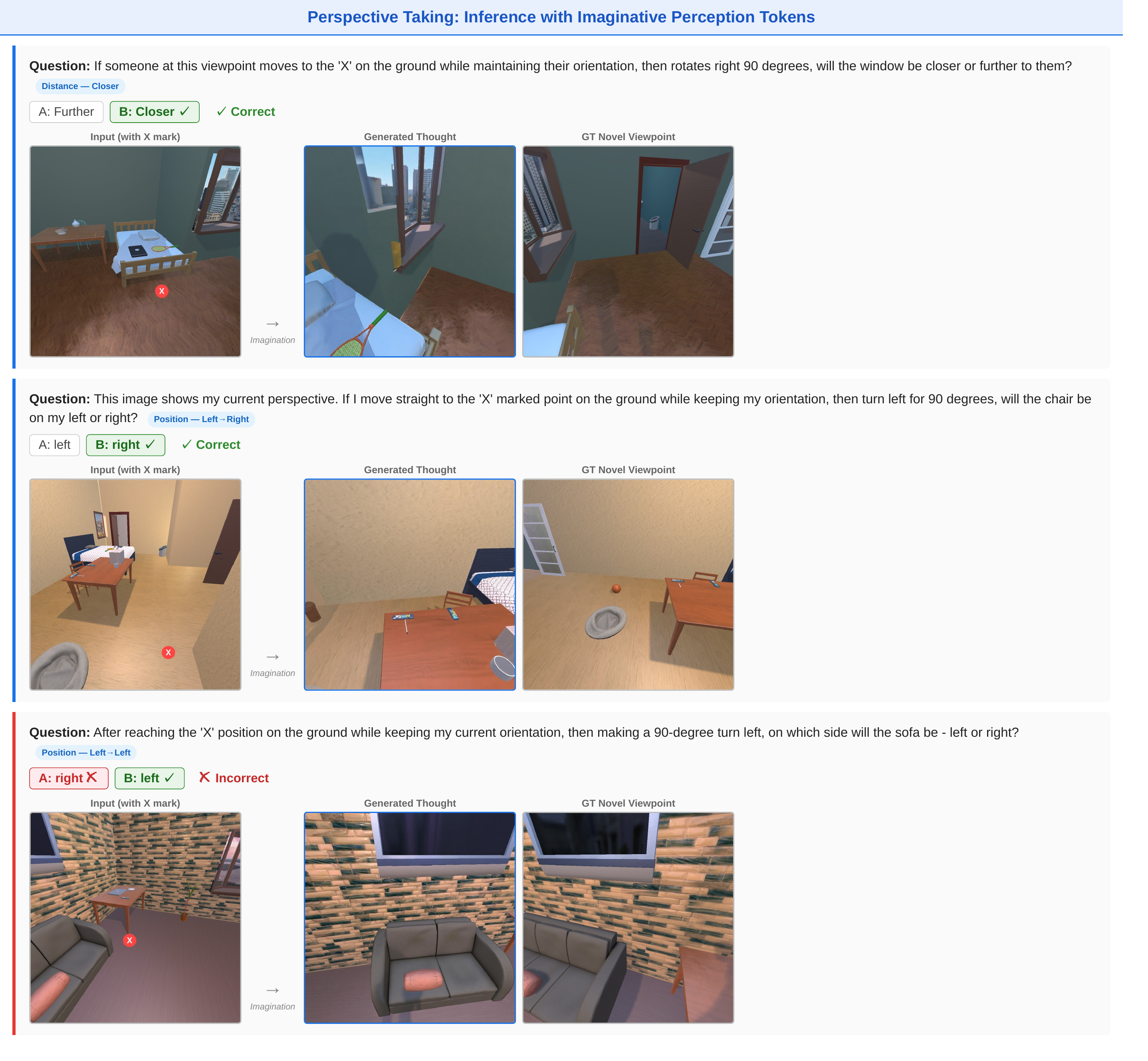}
\caption{\textbf{Perspective taking with imaginative perception tokens.} The model receives an input view with an ``X'' mark on the ground, imagines the scene from the target viewpoint, and predicts spatial relationships. Generated thoughts are compared against ground-truth novel viewpoints. Correct answers are highlighted in green; incorrect predictions in red.}
\label{fig:pet_viz}
\end{figure}

\paragraph{Inference with imaginative perception.}
Figure~\ref{fig:pet_viz} shows examples of perspective taking inference with imaginative perception tokens.
The model receives a first-person view with an ``X'' mark indicating the target position, generates an imagined novel viewpoint as a visual thought, and predicts whether an object is closer/further or on the left/right.
As with path tracing, the generated visual thoughts are not pixel-perfect but capture the essential spatial layout.
The first two rows show correct predictions where the model successfully imagines the scene from the new viewpoint.
The third row shows a failure case where the model incorrectly predicts the relative position.

\begin{figure}[]
\centering
\includegraphics[width=\linewidth]{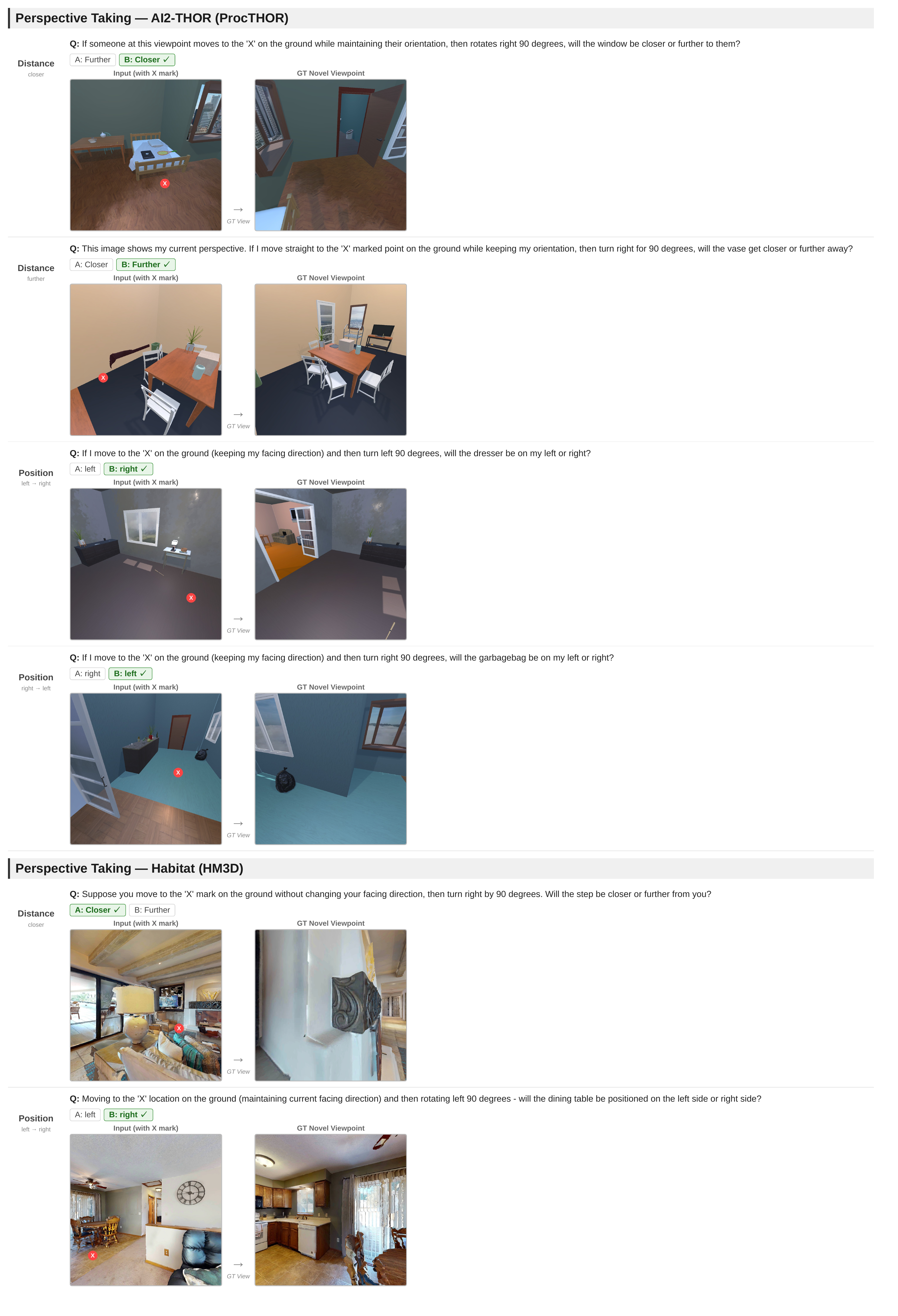}
\caption{\textbf{Perspective taking dataset examples.} Examples from AI2-THOR/ProcTHOR across the four sub-categories. Each example shows the input view with ``X'' mark and the ground-truth novel viewpoint. The correct answer is highlighted in green.}
\label{fig:pet_dataset_examples}
\end{figure}

\paragraph{Dataset examples.}
Figure~\ref{fig:pet_dataset_examples} shows representative examples from the AI2-THOR/ProcTHOR perspective taking dataset across all four sub-categories: distance change (closer/further) and relative position (left$\to$right / right$\to$left).
Each example shows the input image with the target position marked by ``X'' and the ground-truth novel viewpoint after moving to ``X'' and turning.

\subsection{Multiview Counting}

\begin{figure}[]
\centering
\includegraphics[width=.95\linewidth]{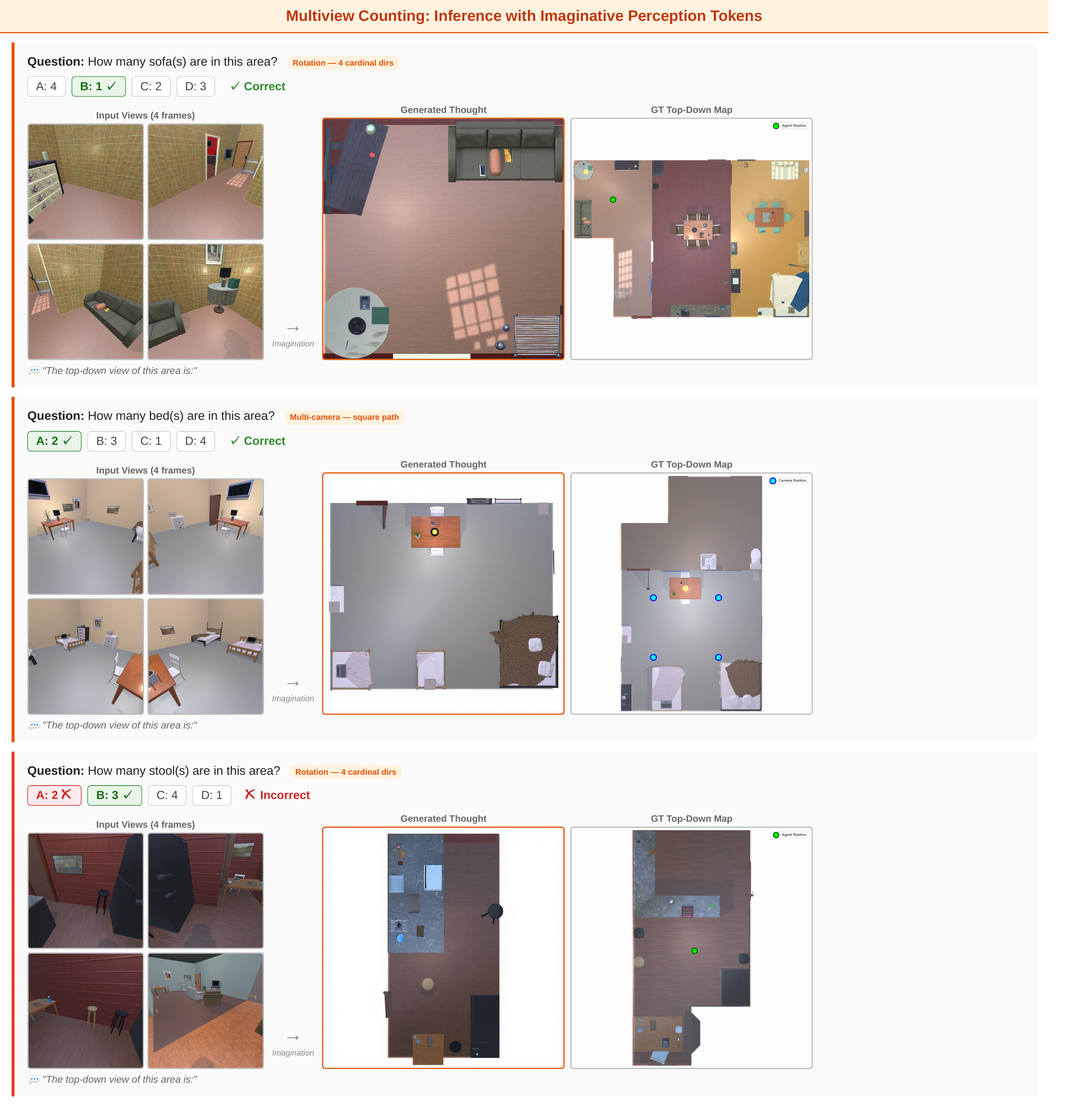}
\caption{\textbf{Multiview counting with imaginative perception tokens.} The model receives four egocentric views, imagines a top-down BEV map, and counts target objects. Generated thoughts are compared against ground-truth top-down maps. Correct answers are highlighted in green; incorrect predictions in red.}
\label{fig:mvc_viz}
\end{figure}

\paragraph{Inference with imaginative perception.}
Figure~\ref{fig:mvc_viz} shows examples of multiview counting inference with imaginative perception tokens.
The model receives four egocentric views from a rotation or multi-camera trajectory, generates a top-down BEV map as its visual thought, and counts the target objects.
The generated top-down maps show that the model learns to synthesize a bird's-eye view from multiple perspectives, capturing the approximate room layout and object placements.
The first two rows show correct predictions; the third row shows a failure case.

\begin{figure}[]
\centering
\includegraphics[height=18cm]{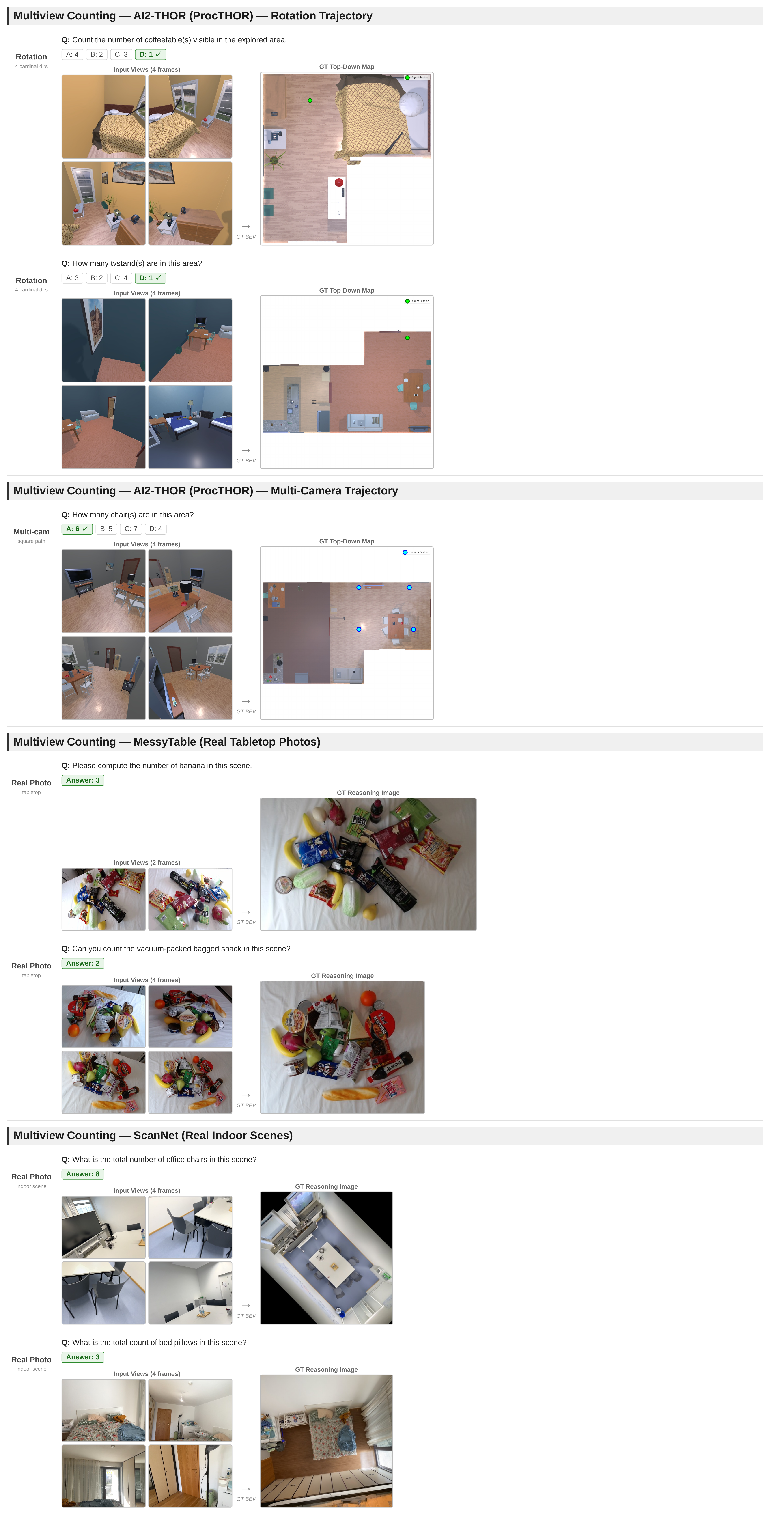}
\caption{\textbf{Multiview counting dataset examples.} Examples from AI2-THOR/ProcTHOR showing both rotation (top) and multi-camera (bottom) trajectories. Each example shows four input views in a $2\times 2$ grid and the ground-truth top-down map. The correct answer is highlighted in green.}
\label{fig:mvc_dataset_examples}
\end{figure}

\paragraph{Dataset examples.}
Figure~\ref{fig:mvc_dataset_examples} shows representative examples from the AI2-THOR/ProcTHOR multiview counting dataset for both trajectory types: rotation (four cardinal directions from a fixed position) and multi-camera (four cameras placed at the corners of a square path).
Each example shows the four input views and the ground-truth top-down map with agent/camera positions annotated.

\section{Imaginative Token Exploration with Different VLMs}
\label{supp_vlms}

Prior to adopting unified models like BAGEL for imaginative perception token generation, we investigated adding discrete imaginative perception tokens directly to the language model vocabulary of state-of-the-art vision-language models. Inspired by Aurora~\cite{bigverdi2024aurora}, we first trained a VQ-VAE from scratch on the intermediate RGB images in our datasets, including novel viewpoint renders, top-down BEV maps, and sideview images. However, the reconstruction quality of these simple VQ-VAEs was insufficient for supervising models on their intermediate token sequences as image outputs.

\begin{figure}[t]
\centering
\setlength{\tabcolsep}{3pt}
\renewcommand{\arraystretch}{1.0}
\begin{tabular}{p{0.23\linewidth}p{0.23\linewidth}p{0.23\linewidth}p{0.23\linewidth}}
\small GT &
\small CB 8K, $f{=}8$ &
\small CB 16K, $f{=}16$ &
\small CB 1K, $f{=}16$ \\[3pt]
\includegraphics[width=3cm,height=3cm,keepaspectratio]{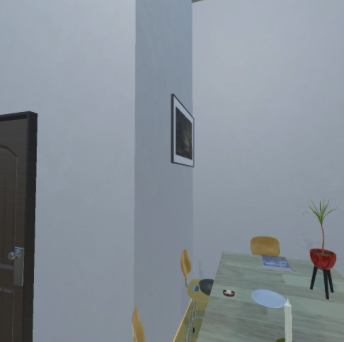} &
\includegraphics[width=3cm,height=3cm,keepaspectratio]{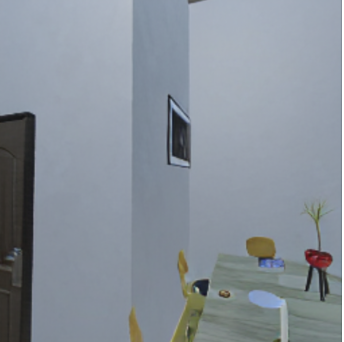} &
\includegraphics[width=3cm,height=3cm,keepaspectratio]{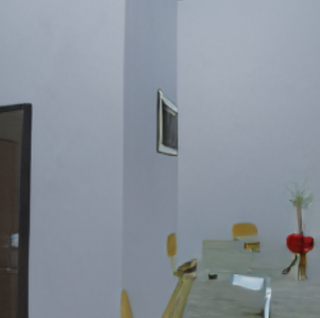} &
\includegraphics[width=3cm,height=3cm,keepaspectratio]{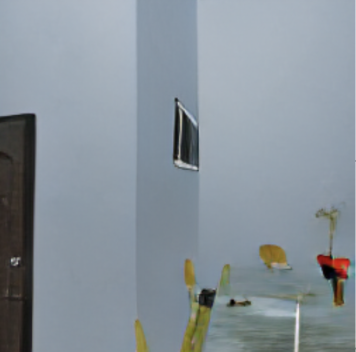} \\[3pt]
\includegraphics[width=3cm,height=3cm,keepaspectratio]{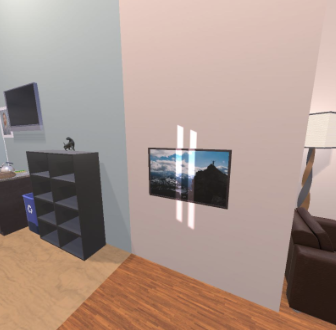} &
\includegraphics[width=3cm,height=3cm,keepaspectratio]{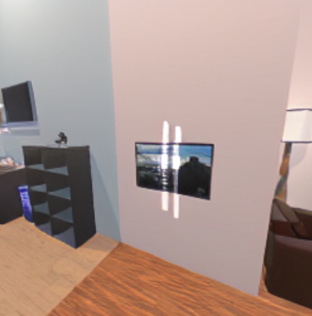} &
\includegraphics[width=3cm,height=3cm,keepaspectratio]{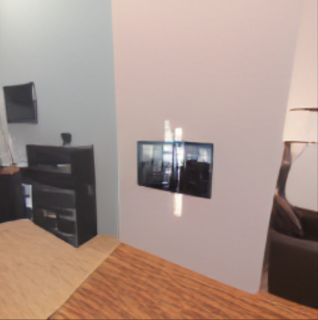} &
\includegraphics[width=3cm,height=3cm,keepaspectratio]{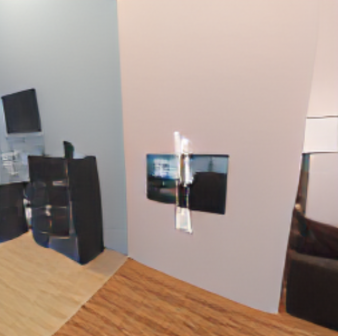} \\
\end{tabular}
\caption{\textbf{VQGAN reconstruction quality across codebook and downsampling settings.} Larger codebooks and smaller $f$ improve fidelity but increase vocabulary size and sequence length respectively.}
\label{fig:vqgan_comparison}
\end{figure}

We therefore switched to off-the-shelf pretrained VQ-GANs~\cite{esser2021taming} with varying configurations. These configurations differ along two axes: codebook size (e.g., 1K, 8K, and 16K entries) and spatial downsampling ratio ($f=8$ vs.\ $f=16$). Each choice involves a tradeoff: a larger codebook improves representational fidelity but inflates the model vocabulary, while a smaller downsampling ratio yields higher reconstruction quality at the cost of a longer token sequence per image, increasing context length. Figure~\ref{fig:vqgan_comparison} illustrates reconstruction quality across these settings.

 We selected Qwen2.5-VL~\cite{wu2025qwenimagetechnicalreport} in two sizes (3B and 7B) as our backbone for discrete token finetuning experiments. We note that the training data at this stage was of lower quality than our final datasets described in main text; these experiments were intended solely to probe whether discrete imaginative perception tokens can serve as a useful intermediate, not to achieve peak performance. We focused on Path Tracing (PT) and Perspective Taking (PET).

For each model we trained three variants: answer-only finetuning, Text CoT, and image chain-of-thought with discrete IPTs, where the VQGAN codebook tokens are appended to the model vocabulary and the model first autoregressively generates the imaginative perception token sequence before predicting the final answer plus a zeroshot baseline. For IPT variants we tested two VQGAN settings that keep sequence length manageable: CB\,16K $f{=}16$ and CB\,1K $f{=}16$. Results are shown in Table~\ref{tab:qwen_discrete}.

IPT consistently outperforms both answer-only finetuning and Text CoT on Path Tracing, with CB\,1K $f{=}16$ yielding the best results for both model sizes (55.0 for 3B, 55.9 for 7B). On Perspective Taking, gains are modest and near the zero-shot baseline, suggesting that lower data quality and the representational limitations of discrete token reconstruction are a bottleneck for this more visually demanding task. A substantial gap remains relative to our final BAGEL-based results, motivating the move to a unified model.

\begin{table}[t]
\centering
\caption{\textbf{Discrete IPT experiments on Qwen2.5-VL.} Accuracy (\%) on Path Tracing (PT) and Perspective Taking (PET). Training data at this stage was of lower quality than the final datasets. ``--'' denotes experiments not conducted.}
\label{tab:qwen_discrete}
\setlength{\tabcolsep}{6pt}
\begin{tabular}{llcc}
\toprule
\textbf{Model} & \textbf{Method} & \textbf{PT} & \textbf{PET} \\
\midrule
\multirow{5}{*}{Qwen2.5-VL 3B}
 & Zero-shot               & 33.0          & 50.0 \\
 & Answer-only             & 48.6          & 50.5 \\
 & Text CoT                & 43.0          & --   \\
 & IPT (CB 16K, $f{=}16$) & 50.5          & 48.5 \\
 & IPT (CB 1K,\phantom{0} $f{=}16$)  & \textbf{55.0} & 50.0 \\
\midrule
\multirow{5}{*}{Qwen2.5-VL 7B}
 & Zero-shot               & 38.5          & 47.2   \\
 & Answer-only             & 37.6          & --   \\
 & Text CoT                & 35.7          & --   \\
 & IPT (CB 16K, $f{=}16$) & 55.0          & --   \\
 & IPT (CB 1K,\phantom{0} $f{=}16$)  & \textbf{55.9} & --   \\
\bottomrule
\end{tabular}
\end{table}
Figure~\ref{fig:qwen_ipt_qual} shows ground-truth imagination images alongside the corresponding IPTs decoded from the Qwen2.5-VL 3B model. The decoded outputs are visually degraded, lacking the spatial structure and object detail present in the ground truth, which helps explain the remaining performance gap and further motivated our switch to continuous latent representations.

\begin{figure}[t]
\centering
\setlength{\tabcolsep}{3pt}
\renewcommand{\arraystretch}{1.0}
\begin{tabular}{p{0.31\linewidth}p{0.31\linewidth}}
\small GT & \small Decoded IPT \\[3pt]
\includegraphics[width=\linewidth,keepaspectratio]{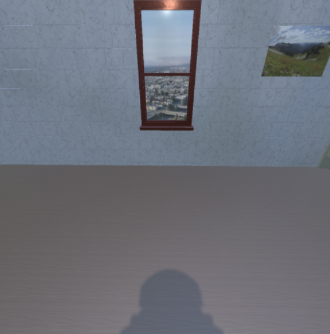} &
\includegraphics[width=\linewidth,keepaspectratio]{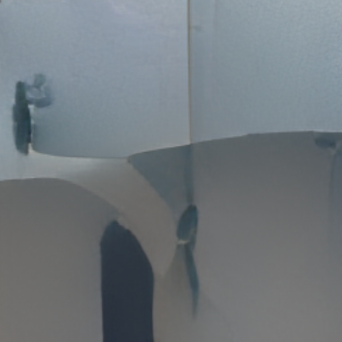} \\
\end{tabular}
\caption{\textbf{Ground-truth vs.\ decoded IPTs from Qwen2.5-VL 3B.} The model-generated imagination tokens decode into visually degraded images that fail to preserve the spatial structure of the ground truth, highlighting the limitations of discrete token generation in non-unified VLMs.}
\label{fig:qwen_ipt_qual}
\end{figure}

We further investigated alternative intermediate image representations. Instead of training the model to generate imaginative perception tokens for RGB thought images, we replaced them with tokens for grayscale images and tokens for pseudo depth maps obtained from the DepthAnything model~\cite{yang2024depth}. The intuition is that simplifying the generation target, from full RGB to grayscale, reduces the difficulty of the token prediction task and may improve spatial reasoning downstream. Results in Table~\ref{tab:qwen_modality} show that switching from RGB to grayscale does boost performance (55.0$\to$59.6 on PT, 50.0$\to$55.5 on PET), while depth tokens perform comparably to RGB. Nevertheless, a substantial gap remains, and Figure~\ref{fig:qwen_gray_qual} shows that the decoded grayscale outputs are still visually degraded, indicating that generation quality rather than representation type is the primary bottleneck.

\begin{table}[t]
\centering
\caption{\textbf{Effect of intermediate image representation on Qwen2.5-VL 3B.} Accuracy (\%) on Path Tracing (PT) and Perspective Taking (PET).}
\label{tab:qwen_modality}
\setlength{\tabcolsep}{6pt}
\begin{tabular}{lcc}
\toprule
\textbf{Method} & \textbf{PT} & \textbf{PET} \\
\midrule
IPT RGB (CB 1K, $f{=}16$)       & 55.0 & 50.0 \\
IPT Grayscale (CB 1K, $f{=}16$) & \textbf{59.6} & \textbf{55.5} \\
IPT Depth (Aurora VQVAE)         & 55.0 & 54.7 \\
\bottomrule
\end{tabular}
\end{table}

Figure~\ref{fig:qwen_gray_qual} shows ground-truth imagination images alongside the corresponding IPTs decoded from the Qwen2.5-VL 3B model. The decoded outputs are visually degraded, lacking the spatial structure and object detail present in the ground truth, which helps explain the remaining performance gap and further motivated our switch to continuous latent representations.

These findings collectively motivated us to move away from discrete token generation in non-unified VLMs and instead adopt a unified model: BAGEL, that natively supports interleaved image understanding and generation through continuous latent representations.

\begin{figure}[t]
\centering
\setlength{\tabcolsep}{3pt}
\renewcommand{\arraystretch}{1.0}
\begin{tabular}{p{0.08\linewidth}p{0.42\linewidth}p{0.42\linewidth}}
 & \centering\small GT & \centering\small Decoded IPT \tabularnewline[3pt]
\small  &
\includegraphics[width=5cm, height=5cm]{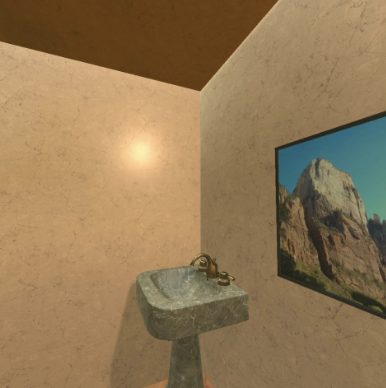} &
\includegraphics[width=5cm, height=5cm]{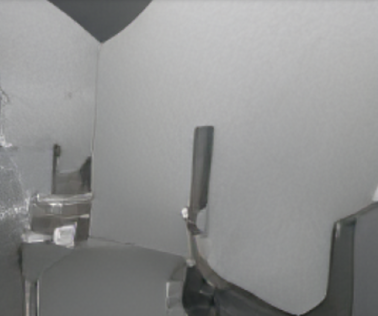} \\[3pt]
\small  &
\includegraphics[width=5cm, height=5cm]{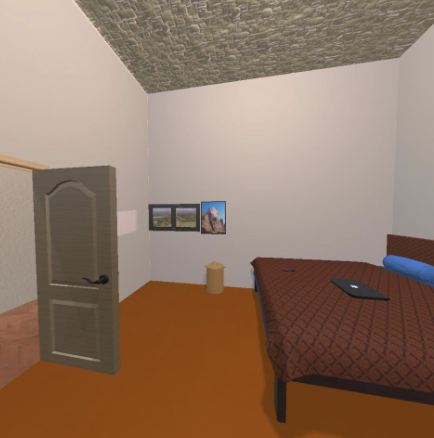} &
\includegraphics[width=5cm, height=5cm]{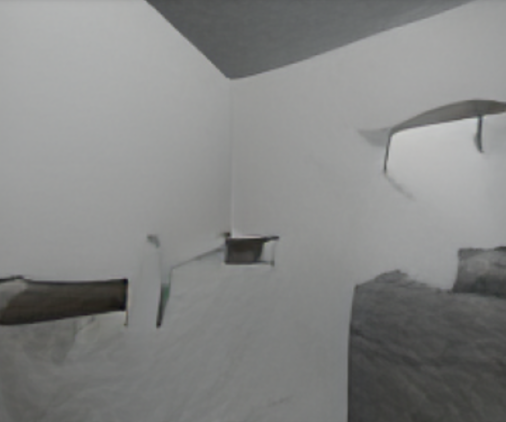} \\
\end{tabular}
\caption{\textbf{Decoded IPTs from Qwen2.5-VL 3B for grayscale representations.} Despite the simpler generation target, grayscale decoded outputs remain visually degraded and fail to preserve spatial structure.}
\label{fig:qwen_gray_qual}
\end{figure}

\end{document}